\documentclass{article}

\usepackage{arxiv}

\usepackage[utf8]{inputenc} 
\usepackage[T1]{fontenc}    
\usepackage{hyperref}       
\usepackage{url}            
\usepackage{booktabs}       
\usepackage{amsfonts}       
\usepackage{nicefrac}       
\usepackage{microtype}      
\usepackage{lipsum}		
\usepackage{graphicx}
\usepackage{natbib}
\usepackage{doi}

\title{Benchmarking LLM Competence on Logical Inference over Probability Operators}

\date{} 					

\author{ {\hspace{1mm}Nayera Hasan}\\
	Haverford College \\
	\And
	 {\hspace{1mm}Jack Greff}\\
	Haverford College \\
    \And
	 {\hspace{1mm}Alvin Grissom~II}\\
	Haverford College \\
}

\renewcommand{\shorttitle}{Benchmarking LLM Competence on Logical Inference over Probability Operators}

\hypersetup{
pdftitle={Benchmarking LLM Competence on Logical Inference over Probability Operators},
pdfauthor={Nayera Hasan, Jack Greff, Alvin Grissom II},
pdfkeywords={llm inference, probability operators, semantic inference, logical inference, llm benchmarks},
}

\usepackage{amsmath}
\usepackage{amssymb}
\usepackage{booktabs}
\usepackage{graphicx}
\usepackage{multirow}
\usepackage{xcolor}
\usepackage[table]{xcolor} 
\usepackage{array}
\usepackage{float}
\usepackage{morefloats}
\usepackage{syllogism}
\usepackage{tabularx}
\usepackage{syllogism}
\usepackage{graphicx}
\usepackage{comment} 
\usepackage{centernot}
\newcommand{\vaff}{\textsc{V.Aff}}
\newcommand{\vneg}{\textsc{V.Neg}}
\newcommand{\iaff}{\textsc{I.Aff}}
\newcommand{\ineg}{\textsc{I.Neg}}
 
\usepackage[disable]{todonotes}  
\newcommand{\nayera}[1]{\todo[inline,color=yellow!35,size=\footnotesize]{\textbf{Nayera:} #1}}

\newcommand{\alvin}[1]{\todo[size=\tiny,color=purple!40]{#1}}
\newcommand{\alvininline}[1]{\todo[inline,size=\tiny,color=purple!40]{#1}}

\makeatletter
\def\@syllog[#1]#2#3#4{%
  \def\@SYLpropA{\ignorespaces#2\unskip}%
  \def\@SYLpropB{\ignorespaces#3\SY@PuncPB\unskip}%
  \def\@SYLpropC{\ignorespaces#4\unskip\nolinebreak\hspace{\SY@SpConEG}\SY@ErgoSign}%
  \settowidth{\SY@LenPA}{\@SYLpropA}
  \settowidth{\SY@LenPB}{\@SYLpropB}
  \settowidth{\SY@LenC}{\@SYLpropC}
  \setlength{\SY@LenLab}{\widthof{#1}}%
  \ifdim\SY@LenLab>\z@
    \addtolength{\SY@LenLab}{\SY@SpLabel}
  \fi
  \setlength{\SY@LenRule}{%
    \fpeval{max(\SY@LenPA,\SY@LenPB,\SY@LenC)}pt}%
  \ifdim\dimexpr\SY@LenRule+\SY@Pre+\SY@LenLab\relax>\dimexpr\columnwidth-\SY@Pre\relax
    \setlength{\SY@LenRule}{\columnwidth-2\SY@Pre-\SY@LenLab}%
  \fi
  \par\vspace{\SY@LenSepA}\noindent 
  \if@SYParam                       
      \hspace{\SY@Pre}#1\hspace{\SY@SpLabel}%
   \else
      \hspace{\SY@Pre}%
  \fi
  \begin{minipage}{\SY@LenRule}
    \raggedright
    \begin{list}{}
      {%
        \setlength{\parsep}{\z@}%
        \setlength{\itemsep}{\z@}%
        \setlength{\leftmargin}{1em}%
        \setlength{\itemindent}{-\leftmargin}%
        \setlength{\labelwidth}{0pt}%
        \setlength{\labelsep}{0pt}%
      }
      \item \@SYLpropA
      \item \@SYLpropB\vspace{\dimexpr-\ht\strutbox+\dp\strutbox}
      \item \rule{\SY@LenRule}{\SY@HiRule}
      \item \@SYLpropC
    \end{list}
  \end{minipage}
  \par\vspace{\SY@LenSepB}
  \@SYRestoreBooleans
}
\makeatletter
 
\begin{document}
 
\maketitle
 


\begin{abstract}
Both expressions of uncertainty and inferences are ubiquitous in natural language, and valid inferences over natural-language expressions of uncertainty are necessary for not only everyday conversations but also for high-stakes domains such as medicine and law. While large language models are increasingly evaluated on logical reasoning tasks, disentangling principled,  symbolic reasoning from clever surface-level pattern matching is fraught with difficulty.
We introduce a benchmark for reasoning over probability operators---inference over sentences with gradable epistemic modals (e.g., \textit{probably}, \textit{might}, \textit{must}) containing 14,320 procedurally-generated English prompts across fifteen inference templates, systematically varying question form, negation strategy, and surface content.
Evaluating 29 models, we find that most show answer biases independent of the logical form, a systematic preference for Yes or No. We summarize this with a competence floor: the worse of a model's accuracy on Yes-correct and No-correct items.
Only 9 of 29 models exceed random chance. We also test variations in question form, verb phrases/activity, and both the gender and origin of names used in the prompts, finding biased responses across every axis.
\end{abstract}
 
\section{Introduction}
\label{sec:intro}
 
Epistemic modals (EMs) express a state of knowledge, uncertainty, credence, or belief, and such modals are ubiquitous in human language~\citep{teller1972epistemic}. In English, epistemic modality (EM) is facilitated largely by modal verbs such as \textit{must}, \textit{might}, \textit{may}, and adverbs such as \textit{probably}, and it pervades both everyday and specialist language.  The sentences \textit{Bob might be home early} and \textit{Bob will possibly be home early}, for example, express that there is some chance, not necessarily a high chance, that Bob will arrive home early, while the epistemic mood of the sentence \textit{Bob must be home early} expresses certainty that he will be.  While modal logics address the subset of cases expressing necessity and possibility---e.g., \textit{must} and \textit{might}---gradable epistemic modality also includes cases expressing \textit{degrees} of certainty, known as \textit{probability operators}~\citep{yalcin2010probability}.  

In our work, we build upon theoretical work that studies inference rules over natural-language sentences containing probability operators~\citep{yalcin2010probability} to benchmark and analyze the capabilities of LLMs in correctly performing zero-shot logical \textit{inference} with a large set of procedurally-generated EM sentences.  EM has received little attention in modern computational linguistics, though earlier work artificial intelligence explored modal and epistemic logics for the integration of uncertain information~\citep{moore1981reasoning}, which is a fundamental problem of AI, broadly defined.  

While there exists philosophy and linguistics literature on the semantics of epistemic modals and inference over them (e.g., \citet{yalcin2010probability,kratzer2012modals}, there is less work on  their analysis in NLP.  \citet{holliday2024conditional} take an important step in this direction, probing LLMs on modal reasoning with \textit{might} and \textit{must}, representable by modal logic.
They find basic errors and logically inconsistent judgments across related inference types, echoing concerns in other areas about the plausible-but-invalid inferences to which LLMs are prone.

\alvininline{Note: Add citations for integrating probability into LLMs in related work.}

This echoes similar work which finds that LLMs' accuracy on syllogisms depends heavily on recognizing specific token patterns, with changes to names, entities, or quantifiers causing predictable performance shifts~\citep{jiang2024peek}, which we also examine.
EM reasoning competence is particularly relevant given that EM inference is central to nearly all high-stakes, natural-language scenarios in which LLMs---in these contexts, often colloquially called AI---are likely to be deployed: a clinical decision-support system reading a note that a treatment ``will probably work'' or a person ``probably has a disease'' must synthesize premises under uncertainty before ranking options; a system triaging legal documents must distinguish between ``might have known'' and ``probably knew'', since the two carry different evidentiary weight; and a financial pipeline that treats ``the market will probably recover'' as equivalent to ``the market will certainly recover'' mischaracterizes risk.
As LLMs are embedded in decision loops in medical, legal, and financial domains, their ability to correctly handle inferences with \textit{probably}, \textit{might}, and \textit{must} becomes a practical concern, not only a theoretical one.
  
Our benchmark covers fifteen templates, encoding thirteen distinct inference patterns (10 valid, 3 invalid) with 14,320 prompts. We vary the wording for logically equivalent inference patterns to check whether the model is changing its inference based on tokens instead of the underlying logic: we use five question forms (\textit{Is it correct that...?}, \textit{Is it true that...?}, \textit{Does it follow that...?}, etc.), multiple negation strategies (prefix, \textit{not}, proposition-level), and surface content (names from seven nationality groups, five activity descriptions). A genuine reasoner will produce the same answer across all of these, irrespective of lexical variation. We evaluate 29 models; most show a fixed bias toward one answer, a systematic preference for \textit{Yes} or \textit{No} that holds across inference types rather than tracking what each inference licenses.

Our primary contributions are: (1)  a dataset and benchmark, tested across a variety of models, for inference over probability operators that varies question form, polarity, and content (verb and noun phrases) for fixed logical forms, disentangling models' answer bias and pattern-matching from principled reasoning~(\S\ref{sec:results}); (2) strong evidence that models answer from a fixed answer bias rather than the logical inference (only 9/29 models exceed a random baseline), and that this finding is negation-independent~(\S\ref{sec:results:overall}); and (3) a comparison of negated sentences and non-negated sentences, revealing accuracy gaps of up to 64 percentage points on semantically identical questions, identifying negation strategy as a confound in benchmark design (\S\ref{sec:results:negstyle}).

\section{Related Work}
\label{sec:related}
 To provide theoretical context for the task and situate our work within prior literature on evaluating LLM reasoning, we review related work.
\nayera{Holliday and Jiang feel closest, but I'm not sure which to make central vs. background. I also don't want this to read as a negation paper. How should I frame this section?}
 
\subsection{Modal and Probabilistic Reasoning in LLMs}
\alvininline{Ideally, we can group this work by theme.}
\nayera{Could you say a bit more about what you'd like grouped by theme here? Want to make sure I  understand what you have in mind?}
 
Prior work on the semantics of epistemic modals provides the theoretical grounding for the inference templates on which we evaluate models.
\citet{kratzer1991modality,kratzer2012modals} establishes the standard quantificational framework; \citet{lassiter2011measurement,lassiter2017graded} proposes the probabilistic alternative in which \textit{probably} $\phi$ entails that $p(\phi)$ is high, \textit{might} $\phi$ that $p(\phi)$ is non-trivial, and \textit{must} $\phi$ that $p(\phi)$ is near-maximal; and
\citet{yalcin2010probability}'s work on probability operators catalogues valid and invalid inference patterns involving epistemic comparatives.
 
\citet{holliday2024conditional} test 29 LLMs on 20 inference patterns using conditionals and the modal auxiliaries \textit{might} and \textit{must}.
They find that all models commit basic fallacies and that even the strongest model in their study (GPT-4, closed source) exhibits logically inconsistent judgments across related patterns (e.g., accepting Modus Tollens with \textit{must} but rejecting it with \textit{might}, which is logically contradictory).
Their templates focus on modal auxiliaries (\textit{might}, \textit{must}) with conditionals, while ours, following semantics work by \citet{yalcin2010probability}, focuses on the adverb \textit{probably}, examining LLM inference patterns specific to graded epistemic reasoning (e.g., distribution over conjunction, conditional-to-comparative, and conjunction fallacy).

\citet{li2025representations} furthermore provide LLMs with short narratives containing factual information and queries the LLMs with prompts containing modal auxiliaries may/might vs. must/have to as well as with attitude verbs (\textit{know}, \textit{believe}, \textit{doubt}).  They note that accuracy is affected by prompt format. Our work is more reductive in its focus on simple templates rather than complex narratives.

\citet{imannezhad2026divergent} examine GPT-5's probabilistic reasoning on conjunction and disjunction fallacies and binary complementarity violations using matched human participant profiles, arguing that GPT-5 appears to have a consistent internal probability model (which our results contravene) that align with quantum-probabilistic models and exceed that of humans.  

They elicit numeric probability ratings under a single fixed prompt format and find the judgments internally coherent. In contrast, we hold the inference fixed and vary the question format, finding that binary judgments shift according to the formulation. Our panel is predominantly open-weight for replicability, with a small number of frontier closed models included as reference points; we do not evaluate the commercial GPT-5 model examined by Imannezhad et al., the closest model in our panel being GPT-5.4-mini. Additionally, rather than compare with human judgments, our work focuses on the inherent correctness of LLM inferences.

 
\citet{macmillanscott2024} tests seven LLMs on cognitive bias tasks from the psychology literature, finding that LLMs are irrational,
but that their irrationality does not reflect human patterns: when models give incorrect answers, they are incorrect in different ways than humans.
They also found significant inconsistency in responses, consistent with our finding that much of what looks like reasoning on these tasks is surface-driven answer behavior rather than proposition-tracking.

\subsection{Surface Sensitivity and Token Bias}
 
LLM performance shifts with surface-form changes that should not affect the answer.  \citet{jiang2024peek} describe this as ``token bias'' and introduce a hypothesis-testing framework using McNemar's test \citep{mcnemar1947} on matched pairs.
They show that on conjunction fallacies and syllogistic problems, changing names, entities, or quantifiers while holding the logic constant produces predictable performance shifts, indicating reliance on superficial patterns rather than reasoning.  Similarly, \citet{binz2023using} find that GPT-3 (which, being closed source, is no longer available) solved canonical cognitive psychology vignettes ``similarly or better than human subjects'' but that ``small perturbations to vignette-based tasks can lead GPT-3 vastly astray.''  Other work highlights that even meaning-preserving \textit{formatting} changes (separator characters, label styles) produce accuracy swings of up to 76 percentage points~\citep{sclar2024quantifying}, echoing similar issues documented in machine translation~\citep{shi-etal-2022-rare}.  Furthermore, LLMs have a bias toward accepting false presuppositions, making them prone to inferences based on implicit misinformation~\citep{ermakova2026confirmation}. \citet{sharma2024sycophancy} directly challenges claims of LLM  and large reasoning model ``reasoning'', demonstrating that the former outperform the latter at low problem complexity but that both collapse at higher problem complexities.  
 
In terms of perturbations, our work---which focuses on a particular, but fundamental, corner of syllogistic reasoning---differs from prior work in several respects.
First, perturbations in prior work change what the problem talks about (names, entities, quantifiers) or how the text looks on the page (separators, label styles). We also do this but also change how the question is asked, swapping the metalinguistic wrapper (\textit{Is it true that...}, \textit{Does it follow that...}) and the way a question's negation is expressed, a dimension of surface sensitivity closer to the logical structure itself.
We also decompose accuracy into orthogonal components (the answer bias and polarity sensitivity), going beyond binary detection of sensitivity to measurement of how much each factor explains. 

\subsection{Response Bias and Acquiescence}
 
\citet{tjuatja2024llms} investigate whether LLMs exhibit human-like response biases (acquiescence, response order, opinion floating).
Testing nine models on 2,578 question pairs, they find that LLMs generally \textit{do not} reflect human-like bias patterns and that RLHF reduces sensitivity to bias-inducing modifications while increasing sensitivity to non-bias perturbations.
\citet{braun2025acquiescence} extended this to 37,975 question variations phrased survey-style (``do you agree...'', ``don't you agree...'') over classification tasks on legal-domain texts in English, German, and Polish, finding that LLMs display a bias toward answering \textit{No} in English, the opposite of human acquiescence bias.
This aligns with our results The direction of the answer bias (yes vs.\ no) is model-dependent rather than universal, with some models strongly yes-biased and others strongly no-biased.

Our baseline\alvin{Instead of ``competence floor'', please say ``baseline''} 
\nayera{Happy to switch my only note is "baseline" is already our word for the overall 0.5 chance level, while the competence floor is the minimum of the accuracies, which we read against that same 0.5 so renaming the floor to "baseline" merges the metric and its chance level into one word do you want to use baseline for both?}
takes the minimum of the two answer-conditioned accuracies rather than their mean because the mean alone obscures the bias: mean accuracy can be high when one answer class is near ceiling and the other near floor.
The benchmark's validity-by-negation structure is what makes this minimum informative, providing an absolute baseline under which a maximally biased LLM scores 0 regardless of the valid/invalid mix.
We defer the full construction to the discussion of accuracy metrics~(\S\ref{sec:results:overall}).
 
A third evaluation framework treats answer-label priors as a calibration problem.
\citet{zhao2021calibrate} show that few-shot classification with language models is skewed by majority-label, recency, and common-token biases, estimated the model's prior from a content-free input; \citet{zhao2021calibrate} divide it out at inference,\alvin{What does `divide it out' mean here?} \nayera{Good question, I went back to the paper from what I can tell, their contextual calibration feeds a content-free input ("N/A") to estimate the model's label prior $\hat{p}_{cf}$, then sets $W=\mathrm{diag}(\hat{p}_{cf})^{-1}$, $b=0$ and predicts $\arg\max(W\hat{p})$, which divides each label's score by that prior so it cancels. this is why I used "divide it out" Would it read more clearly if I changed it to "normalize each label's prediction by the content-free prior, so the prior cancels"? Happy to adjust if you'd phrase it differently}  with large accuracy gains; \citet{fei2023mitigating} refined the prior estimate with in-domain words, and \citet{zhou2024batch} with the mean prediction over a test batch.
\citet{holtzman2021surface} traced related distortions to probability mass\alvin{of accuracy?} \nayera{im not sure it's accuracy here in Holtzman et al the probability mass splits across surface strings for the same answer (like "computer" vs "PC") which lowers any one correct form. want me to add "(the model's predicted probability)" on first mention so it doesn't read as accuracy?} splitting across surface forms of the same answer.
These results were established on models from 2021 to 2024; our panel suggests the phenomenon has not aged out, with current instruction-tuned and reasoning models still carrying extreme answer priors.
Our answer bias is a label bias on a Yes/No space, and this literature suggests such priors are often removable by calibration.
Where that work treats the prior as a nuisance to be removed so that accuracy improves, we instead report it directly through the competence floor, which remains diagnostic whether or not the prior could be calibrated away.

\subsection{Negation Processing}
 
\citet{truong2023language} evaluate GPT-neo, GPT-3, and InstructGPT on six negation benchmarks, finding three key limitations: insensitivity to the presence of negation,\alvin{So they ignore it?} \nayera{pretty much, yeah. i checked Truong et al again  their "insensitivity to the presence of negation" means the model gives basically the same answer whether or not negation is there, like it doesn't register it. want me to just say "models often answer the same way with or without negation" so it's clearer?} inability to capture its lexical semantics, and failure to reason under negation.
They also find that negation shows flat or inverse scaling (larger models are \textit{more} insensitive), a pattern alleviated only by instruction fine-tuning.
\citet{garciaferrero2023} introduced a large negation benchmark of approximately 400,000 sentences, confirming that LLMs rely on superficial cues and that fine-tuning improves but does not fully resolve the problem.
\citet{so2025thunder} establish a typology distinguishing morphological negation (\textit{un-}, \textit{in-}) from syntactic negation (\textit{not}), which directly corresponds to the contrast we test.
\citet{elkins2026prohibitions} audit negation sensitivity in ethical decisions, finding that open-source models endorse prohibited actions\alvin{What's a prohibited action?} \nayera{it's their term for an action a rule says must not be taken, like a contraindicated treatment or a loan applicant who must not be approved. the model fails when it says go ahead anyway} 77\% of the time under simple negation and 100\% under compound negation.

Our benchmark contains prompts with no negation at all, some expecting Yes and some expecting No, so the core contrast does not depend on negation.
On top of that, we implement negation in several ways, asking whether a model that accepts a valid inference (or rejects an invalid one) in the affirmative form can still do so when the question is negated.
Implementing it became a source of findings in its own right: writing the same negation as a prefix or with the word \textit{not} produces dramatically different accuracy on semantically identical content (\S\ref{sec:results:negstyle}).

\section{Benchmark Design}
\label{sec:benchmark}
 In this section, we describe the overall design of our benchmarks, including the inference templates and their logical forms.  
 
\subsection{Inference Templates}
\label{sec:benchmark:templates}
 Our work is theoretically motivated by prior work on EM semantics which characterizes valid inferences. In particular, we create prompts based on probability operators analyzed by \citet{yalcin2010probability}, crafted as binary questions with a known correct response.
 
 Thirteen inference templates compose the benchmark of EM expressions; for exposition, we group the templates into three groups based on validity and complexity.
Eleven inference rules have a single template; two (distribution over conjunction and conjunctivitis; \S\ref{sec:analysis:conjvariants}) each have two surface-form templates.  Altogether, we have fifteen.

\alvininline{Terms (as below) needs to be defined and explained before used in this way.} 
\nayera{so do we move the table of the templates above this? I don't want us to duplicate the information \alvin{Typically, we want to reference the appropriate section later.}}
Table~\ref{tab:templates} shows each pattern with a representative prompt from the dataset with the logical form described in terms of probability and modal operators for necessity ($\square$) and possibility ($\lozenge$).  We use $\succeq$ to denote ``is at least as likely as``, i.e., $A \succeq B \iff p(A) \ge p(B)$.  For brevity, we also use $Pr(A)$ to denote ``probably A'' and $p(A)$ for a numerical probability, i.e., $Pr(A) \iff p(A) \ge 0.5$ under typical probability assumptions.
\subsubsection{Template Descriptions}
 
\nayera{Do we actually want to group the templates at all? If yes, is validity and complexity the right grouping, or something else like difficulty?\alvininline{I think validity is reasonable, but I only see two groups in the table.}}

\nayera{good catch. the table was only meant to list the inferences with their schema and example, so it just split valid vs invalid, but the prose promises three groups, so they don't line up. i think the cleanest fix is to make the table the single list, reorder it into three labeled bands (simple valid, multi-step valid, invalid), and trim the Group paragraphs so they explain what separates the tiers instead of re-listing each template, since right now we say everything twice. does that work, or would you rather i just keep the prose and add the third divider to the table?}

\paragraph{Simple valid inferences (5 templates)}
These test straightforward relationships between epistemic operators.

\begin{itemize}
\item[] \textit{Probably} to \textit{Might}: if $\phi$ is probable, then $\phi$ is possible.
\item[] \textit{Must} to \textit{Probably}: if $\phi$ is necessary, then $\phi$ is probable.
\item[] \textit{Probably} to \textit{Not Probably Not}: if $\phi$ is probable, then it is not probable that $\neg\phi$.
\item[] \textit{Positive Form Transfer} (PFT) and \textit{Complement Transfer} (CT) test probability preservation under reformulation. 
\syllog[Positive Form Transfer] 
{$\psi$ is at least as likely as $\phi$} %
{probably $\phi$} %
{probably $\psi$}

\syllog[Complement Transfer] 
{$\psi$ is at least as likely as $\phi$} %
{$\phi$ is at least as likely as $\lnot\phi$} %
{$\psi$ is at least as likely as $\lnot\phi$}

\end{itemize}
 
\paragraph{Multi-step valid inferences (5 templates)}
Chancy Modus Ponens and its contrapositive variant Chancy Modus Tollens require combining premises or comparing probabilities.
\begin{itemize}

\item[]  \syllog[Chancy Modus Ponens] 
{if $\phi$ then $\psi$} %
{probably $\phi$} %
{probably $\psi$}

\syllog[Chancy Modus Tollens] 
{if $\phi$ then $\psi$} %
{$\lnot$ probably $\psi$} %
{$\lnot$ probably $\phi$}

\item[] \textit{Distribution over conjunction}: if probably ($\phi \land \psi$), then probably $\phi$ and probably $\psi$.
\item[] \textit{Conditional to Comparative}: if probably ($\phi \to \psi$), then $\psi$ is at least as probable as $\phi$.
\item[]\textit{Chancy Disjunction Introduction}: if probably $\phi$, then probably ($\phi \lor \psi$).
 \end{itemize}
 
\paragraph{Invalid inferences (3 templates)}
These measure correct rejection of fallacious reasoning.
\begin{itemize}
\item[] \textit{Conjunctivitis}~\citep{kyburg1970conjunctivitis} is a fascinating conjunction fallacy discussed in the philosophical literature.  The typical logical assumption of closure under conjunction does not hold for probability operators.  Closure under conjunction holds when the truth of $\psi$ and $\phi$ independently implies the truth of $\psi \land \phi$.  It holds for non EM predicates but not for the probability operator.  I.e., 
$\text{probably}(\psi), \text{probably}(\phi) \centernot\implies \text{probably}(\psi\land\phi)$, a fact that can be illustrated by the lottery paradox, whereby we have a collection of propositions of probability at least $0.5$ but less than $1$, whose conjunction probability necessarily decreases with each new proposition.  Since this is a known fallacy discussed in the literature, one might expect that LLMs would be less prone to fall for it.

\item[] \textit{Might} to \textit{Probably} inference: possibility does not entail probability. 
$\Diamond \phi \centernot\implies \Box\phi$.
\item[] \textit{Probably} to \textit{Certain}: probability does not entail certainty. 

probably $\phi \centernot\implies \Box\phi$.
 \end{itemize}

\begin{table*}[!htb]
\caption{Inference templates with formal schema and example prompts (\textsc{truth} question form, affirmative; the negated condition replaces \textit{Is it true that} with \textit{Is it false that} and flips the correct answer). In the affirmative examples shown the correct answer is \textit{Yes} for the valid templates and \textit{No} for the invalid ones, and negation reverses both. $Pr(\phi)$: \textit{probable}; $M(\phi)$: \textit{possible}; $\Box\phi$: \textit{certain}; $\phi \succcurlyeq \psi$: \textit{at least as likely}. Names and activities vary.}
\label{tab:templates}
\footnotesize
\centering
\setlength{\tabcolsep}{4pt}
\renewcommand{\arraystretch}{1.15}
\begin{tabular}{|p{2.4cm}|p{1.3cm}|p{2.9cm}|p{5.9cm}|}
\hline
\textbf{Template} & \textbf{Validity} & \textbf{Schema} & \textbf{Example prompt} \\
\hline
Probably to Might & Valid & $Pr(\phi)$ $\Rightarrow M(\phi)$ & Savir is probably going to be at the party. Is it true that Savir might be at the party? \\ 
\hline
Must to Probably & Valid & $\Box\phi$ $\Rightarrow Pr(\phi)$ & Savir must be at the party. Is it true that it is probable that Savir will be at the party? \\ 
\hline
Chancy Modus Ponens & Valid & $[\phi \Rightarrow \psi, Pr(\phi)] \Rightarrow Pr(\psi)$ & If Savir is going to be at the party, then Ashwin is going to be at the party. Savir is probably going to be at the party. Is it true that Ashwin is probably going to be at the party? \\
\hline
Chancy Modus Tollens & Valid & $Pr(\phi{\to}\psi),$ $Pr(\neg\psi)$ $\Rightarrow Pr(\neg\phi)$ & If Savir is going to be at the party, then Ashwin is going to be at the party. Ashwin is probably not going to be at the party. Is it true that Savir is probably not going to be at the party? \\ 
\hline
Distribution over Conjunction & Valid & $Pr(\phi \land \psi)$ $\Rightarrow Pr(\phi)$ & It is probable that Savir and Ashwin will be at the party. Is it true that it is probable that Savir will be at the party? \\ 
\hline
Conditional to Comparative & Valid & $Pr(\phi{\to}\psi)$ $\Rightarrow \psi \succcurlyeq \phi$ & If Savir is going to be at the party, then Ashwin is going to be at the party. Is it true that Ashwin is at least as likely as Savir to be at the party? \\ 
\hline
Chancy Disjunction Introduction & Valid & $Pr(\phi)$ $\Rightarrow Pr(\phi \lor \psi)$ & Savir is probably going to attend the event. Is it true that it is probable that Savir will attend the event or give a presentation? \\ 
\hline
Positive Form Transfer & Valid & $\phi \succcurlyeq \psi,$ $Pr(\psi) \Rightarrow Pr(\phi)$ & Ashwin is at least as likely as Savir to be at the party. Savir is probably going to be at the party. Is it true that Ashwin is probably going to be at the party? \\ 
\hline
Probably to not probably not & Valid & $Pr(\phi)$ $\Rightarrow \neg Pr(\neg\phi)$ & Savir is probably going to be at the party. Is it true that it is not probable that Savir will not be at the party? \\ 
\hline
Complement Transfer & Valid & $\phi \succcurlyeq \psi,$ $\psi \succcurlyeq \neg\psi$ $\Rightarrow \phi \succcurlyeq \lnot\phi$ & Ashwin is at least as likely as Savir to be at the party. Savir is at least as likely to be at the party as to not be at the party. Is it true that Ashwin is at least as likely to be at the party as to not be at the party? \\
\hline
Conjunctivitis & Invalid & $Pr(\phi), Pr(\psi)$ $\centernot\implies Pr(\phi \land \psi)$ & It is probable that Savir will be at the party. It is probable that Ashwin will be at the party. Is it true that it is probable that Savir and Ashwin will be at the party? \\
\hline
Might to Probably & Invalid & $M(\phi)$ $\centernot\implies Pr(\phi)$ & Savir might be at the party. Is it true that it is probable that Savir will be at the party? \\
\hline
Probably to Certain & Invalid & $Pr(\phi)$ $\centernot\implies \Box\phi$ & Savir is probably going to be at the party. Is it true that it is certain that Savir will be at the party? \\
\hline
\end{tabular}
\end{table*}

\subsection{Question Form}
\label{sec:benchmark:framing}
\begin{table*}[t!]
\caption{The five English question forms and their negated versions, plus the two \textit{not}-prefixed forms, shown on \textit{Must} to \textit{Probably} (premise: ``Savir must be at the party''). The repeated conclusion ``it is probable that Savir will be at the party'' is abbreviated as \dots. Correct answers flip with polarity: Yes affirmative, No negated.}
\label{tab:frames}
\small
\centering
\begin{tabular}{@{}lll@{}}
\toprule
\textbf{Question form} & \textbf{Affirmative (Yes)} & \textbf{Negated (No)} \\
\midrule
\textsc{truth}   & Is it true that \dots?              & Is it false that \dots? \\
\textsc{correct} & Is it correct that \dots?           & Is it incorrect that \dots? \\
\textsc{valid}   & Is it valid to conclude that \dots? & Is it invalid to conclude that \dots? \\
\textsc{follow}  & Does it follow that \dots?          & Does it not follow that \dots? \\
\textsc{direct}  & Is it probable that Savir will be at the party? & Is it not probable that \dots \\
\midrule
\textsc{not-correct} & ---                            & Is it not correct that \dots? \\
\textsc{not-valid}   & ---                            & Is it not valid to conclude that \dots? \\
\bottomrule
\end{tabular}

\vspace{2pt}
{\footnotesize \textsc{direct} carries no metalinguistic wrapper, negating the proposition itself (\textit{probable}  to  \textit{not probable}).}
\end{table*}

We present questions in both \textit{affirmative} and \textit{negated} forms to test consistency under negation.
In the affirmative condition, we ask whether the conclusion holds (correct answer: \textit{Yes} for valid; \textit{No} for invalid).
In the negated condition, we ask whether the same conclusion does not hold (correct answer: \textit{No} for valid, \textit{Yes} for invalid).
 
Five metalinguistic question forms vary how the question is asked.
\textsc{correct}, \textsc{truth}, and \textsc{valid} negate by switching to the negative adjective (correct  to  incorrect, true  to  false, valid  to  invalid).
\textsc{follow} inserts the word \textit{not}  changing, for example, \textit{does it follow}  to  \textit{does it not follow}).
\textsc{direct} negates the proposition itself.
 
An additional pair of question forms, \textit{Is it not correct that\,\dots?} and \textit{Is it not valid to conclude that\,\dots?}, replaces the negative adjective with the word \textit{not} on the same adjective, enabling direct comparison of negation strategies on identical semantic content (\S\ref{sec:results:negstyle}). In our analysis, we discuss the perlocutionary force of such questions in light of known issues of LLM sycophancy (\S\ref{sec:results:negstyle}).  Specifically, for our question templates, introducing negation can pragmatically encourage the model to agree with the questioner.  Asking, \textit{Is it not probable that $\phi$?} can, under a common interpretation, sound like the answer \text{should} be the \textit{Yes}.  We introduce clauses such as \textit{Is it true that} for affirmative and \textit{Is it false that} for the negation to enforce some neutral symmetry in the question surface form without such pragmatic force skewing the results.

\subsection{Demographic Bias Controls}
\label{sec:benchmark:controls}
 
Natural language models are known to be prone to spurious biases.  To control for this, names are drawn from seven nationality groups (Indian, Russian, Japanese, African, German, French, American) plus abstract letter variables, balanced by gender (common names for men or women).
Activities include five semantically similar scenarios.
These function as robustness checks: a model that reasons about the logic should show minimal accuracy variation across these controls.

\subsection{Models Evaluated}
\label{sec:benchmark:models}
 
We evaluate 29 models in English spanning five size-graded families (Gemma~3: 270M--27B; Gemma~4: E2B--31B; Qwen~3: 0.6B--32B; DeepSeek-R1: 7B--32B; Llama~3: 3B, 8B) plus frontier models (Claude Opus~4.7, Claude Sonnet~4.6, GPT-oss 120B, GPT-oss 20, GPT-5.4-mini), Granite~3.2:8B, Mistral:7b, and the Arabic-oriented Aya-Expanse 8B and 32B.

 
We show all models the same 80 surface prompts per template, and we run each model once per prompt at temperature $\tau=0$ All main analyses pool across templates and question forms.
 
\paragraph{Prompting setup}
We query open-weight models zero-shot, and the first answer token is parsed as \textit{Yes}, \textit{No}, or classified as \textit{Uncertain} otherwise.
\alvin{Do we instruct the model to answer YES or NO specifically?}
Two choices bear on interpretation.
First, to keep the evaluation strictly zero-shot, we disable chain-of-thought wherever the model permits it: Qwen~3 and DeepSeek-R1 are run with thinking turned off, and GPT-oss is set to its lowest available reasoning level.
The reported numbers therefore reflect direct-answer behavior.
Second, Qwen~3:4B is queried with a constrained Yes/No output format, because unconstrained generation produced a non-compliance rate as high as 96\%.
Constraining the output still lets the model choose either answer, so it does not manufacture the answer bias we report.

\subsection{Metrics}
\label{sec:benchmark:metrics}

We score each model on a 2$\times$2 grid, crossing inference validity (valid/invalid) with question negation (affirmative/negated); we refer to its four cells throughout. This crossing determines the correct answer in each cell.
\alvininline{We need to specify why always guessing Yes or No leads to high baselines.}
We use abbreviations to refer to the accuracy for each intersection of conditions. V.AFF, for example, refers to a \textit{valid} template with a question phrased in the affirmative, while I.NEG refers to an \textit{invalid} template with a negated question.
The correct answer is \textit{Yes} in two cells (\vaff{} and \ineg{}) and \textit{No} in the other two (\vneg{} and \iaff{}).

Overall accuracy is the fraction of correct responses (with uncertain counted as incorrect), computed over whatever set of prompts a table covers. Since valid templates outnumber invalid ones, weighting the four validity$\times$negation cells equally instead pulls every strong model toward 0.5, by up to about 9 points; supplementary Table~\ref{tab:pooling} reports accuracy under five weighting schemes so the sensitivity to this choice is visible. 

To separate the model's surface-level bias from principled proposition-tracking, we measure the yes/no bias.  We do this by averaging the model accuracy, respectively, on questions for which the answers should be affirmative (\textit{Yes}) and on questions for which the answers should be negative (\textit{No}).  For each, we have two cases, the valid and invalid conditions.\footnote{These are true positive rates, but we avoid using the term since ``'positive'' is confusing when discussing yes/no questions that can be negated.}
Let $\text{Acc(Yes)} = \frac{1}{2}(\text{\vaff} + \text{\ineg})$ and $\text{Acc(No)} = \frac{1}{2}(\text{\vneg} + \text{\iaff})$\alvin{Is V.AFF actually Acc(V.AFF)}.

The signed gap is the bias toward \textit{Yes} or \textit{No}s:
\begin{equation}
\text{Bias} = \text{Acc(Yes)} - \text{Acc(No)},
\end{equation}
If a model always guesses \textit{Yes}, the bias will be 1; if it always guesses \textit{No}, it will be -1; and a random baseline has a bias of 0\footnote{We have more valid templates than invalid ones; averaging them prevents in this way prevents the valid category from dominating the bias and associated metrics.  However, the average still provides useful information.}.

We also track polarity sensitivity (PS), the model's accuracy drop under negation:
\begin{align}
\text{PS} &= \frac{(\text{\vaff} + \text{\iaff}) - (\text{\vneg} + \text{\ineg})}{2} \\
&= \frac{\left[ \text{Acc(non-negated questions)} - \text{Acc(negated questions)} \right ]}{2}
\end{align}
Bias is positive when the model favors \textit{Yes}; PS is positive when negation hurts.  Bias is an indirect way of measuring a model's \textit{competence}, i.e., a model's actual reasoning ability. To measure competence, we take the worse of the two accuracy measurements, the \emph{floor}:
\begin{equation}
\text{Floor} = \min\!\big(\text{Acc(Yes)},\, \text{Acc(No)}\big).
\end{equation}
Intuitively, this related metric concisely describes how poorly the model performs on its worst-performing question polarity (those with a correct answer of either \textit{Yes} or \textit{No}).  This is motivated by our observation that models tend to be extremely biased for one answer or the other, irrespective of the internal logic.
A constant responder scores 0 on the floor, a coin-flipper 0.5, and a competent reasoner near 1, so the floor has an absolute baseline that overall accuracy lacks.

\section{Results}
\label{sec:results}
We now describe the EM benchmarking results.  While some of the differences in accuracy across conditions may seem small (but many are not), given the query rate of LLMs, seemingly small differences in the number of errors can have huge impacts in the aggregate.  If, for example, an LLM is queried 1 billion times, a $0.3\%$ increase in errors is equivalent to $3$ million more.  Currently, ChatGPT alone receives approximately 2.5 billion queries per day, and GPT-4o received approximately 774 billion queries in 2025, which is still only a fraction of Google's queries, which now integrate LLMs~\citep{search-stats}.
 
\subsection{Overall Accuracy}
\label{sec:results:overall}
 
First, we report the aggregate accuracy across all EM prompts for a bird's eye view of respective model performance as a prelude to more specific analysis. 
Table~\ref{tab:en_full} presents the four cells for all 29 models, using the metrics defined in \S\ref{sec:benchmark:metrics}.

Overall accuracy ranges from 20.2\% (DeepSeek-R1:7B) to 81.8\% (Gemma4:31B), but this obscures the strategies (such as they are) being used.
Qwen3:0.6B scores 54.8\% overall by nearly always responding \textit{Yes} (\vaff{}${}=0.997$, \vneg{}${}=0.160$, \iaff{}${}=0.030$, \ineg{}${}=0.878$), while Llama3.1:8B is balanced (bias only $+$0.03), yet every cell performs near the random baseline, suggesting that the predictions are themselves arbitrary and disconnected from the prompt.

Uncertain responses are rare for most models, below 2\%, but high for a handful (e.g., DeepSeek-R1:7b is uncertain 61.5\% of the time).
Uncertain rates are relatively flat across conditions (expected-Yes 3.2\% vs.\ expected-No 3.4\%), so counting them as incorrect does not unduly affect the competence floor. We report rates in Table~\ref{tab:en_full}.
 
\begin{table*}[t]
\caption{English models, the four cells, sorted by floor; the faded rule separates the nine models that clear the 0.5 floor from those that do not. The four columns \vaff, \vneg, \iaff, \ineg  are per-cell accuracies, so e.g.\ \vaff${}={}$Acc(V.Aff). \textbf{Acc} is overall accuracy. \textbf{Unc}ertain responses count as incorrect. Bias${}=\text{Acc(Yes)}-\text{Acc(No)}$ (signed; positive${}={}$yes-biased). Floor${}=\min(\text{Acc(Yes)},\text{Acc(No)})$ and has a random baseline of 0.5; floors above random chance are in \textbf{bold}. Per-condition accuracies below chance are \underline{underlined}.}
\label{tab:en_full}
\footnotesize
\centering
\begin{tabular}{@{}lrrccccrrc@{}}
\toprule
\textbf{Model} & \textbf{Acc} & \textbf{Unc} & \vaff{} & \vneg{} & \iaff{} & \ineg{} & \textbf{Bias} & \textbf{PS} & \textbf{Floor} \\
\midrule
gemma4:31b & 0.818 & 0.0\% & 0.977 & 0.761 & 0.980 & \underline{0.372} & $-$0.20 & $+$0.41 & \textbf{0.675} \\
Sonnet 4.6 & 0.751 & 0.0\% & 0.994 & 0.636 & 0.648 & \underline{0.491} & $+$0.10 & $+$0.26 & \textbf{0.642} \\
gemma4:12b & 0.764 & 0.0\% & 0.967 & 0.754 & 0.735 & \underline{0.261} & $-$0.13 & $+$0.34 & \textbf{0.614} \\
qwen3:32b & 0.718 & 0.0\% & 0.813 & 0.708 & 0.827 & \underline{0.374} & $-$0.17 & $+$0.28 & \textbf{0.594} \\
gemma3:27b & 0.648 & 1.0\% & 0.913 & \underline{0.470} & 0.712 & \underline{0.334} & $+$0.03 & $+$0.41 & \textbf{0.591} \\
gpt-oss:20 & 0.737 & 0.0\% & 0.815 & 0.740 & 0.894 & \underline{0.356} & $-$0.23 & $+$0.31 & \textbf{0.586} \\
Opus 4.7 & 0.756 & 0.0\% & 0.892 & 0.762 & 0.870 & \underline{0.248} & $-$0.25 & $+$0.38 & \textbf{0.570} \\
gpt-oss:120b & 0.724 & 0.0\% & 0.727 & 0.808 & 0.889 & \underline{0.326} & $-$0.32 & $+$0.24 & \textbf{0.526} \\
gemma3:12b & 0.615 & 0.0\% & 0.812 & 0.620 & \underline{0.449} & \underline{0.224} & $-$0.02 & $+$0.21 & \textbf{0.518} \\
\arrayrulecolor{black!30}\midrule\arrayrulecolor{black}
gemma4:e4b & 0.608 & 0.0\% & 0.652 & 0.655 & 0.674 & \underline{0.294} & $-$0.19 & $+$0.19 & 0.473 \\
llama3.1:8b & 0.463 & 14.3\% & \underline{0.445} & \underline{0.435} & 0.504 & 0.544 & $+$0.03 & $-$0.01 & 0.469 \\
gpt-5.4-mini & 0.656 & 0.0\% & 0.671 & 0.710 & 0.887 & \underline{0.239} & $-$0.34 & $+$0.30 & 0.455 \\
aya-expanse:32b & 0.624 & 0.1\% & 0.905 & 0.522 & \underline{0.311} & \underline{0.440} & $+$0.26 & $+$0.13 & 0.416 \\
gemma3:270m & 0.500 & 0.0\% & 0.640 & \underline{0.369} & \underline{0.456} & 0.512 & $+$0.16 & $+$0.11 & 0.413 \\
gemma3:4b & 0.529 & 0.0\% & 0.603 & 0.524 & \underline{0.283} & 0.589 & $+$0.19 & $-$0.11 & 0.403 \\
qwen3:8b & 0.599 & 0.0\% & 0.925 & \underline{0.410} & \underline{0.386} & \underline{0.424} & $+$0.28 & $+$0.24 & 0.398 \\
qwen3:14b & 0.605 & 0.0\% & \underline{0.472} & 0.813 & 0.729 & \underline{0.281} & $-$0.39 & $+$0.05 & 0.377 \\
aya-expanse:8b & 0.586 & 0.6\% & 0.859 & \underline{0.467} & \underline{0.224} & 0.517 & $+$0.34 & $+$0.05 & 0.346 \\
mistral:7b & 0.535 & 0.0\% & 0.836 & \underline{0.287} & \underline{0.310} & 0.604 & $+$0.42 & $+$0.13 & 0.298 \\
granite3.2:8b & 0.607 & 0.0\% & \underline{0.406} & 0.863 & 0.909 & \underline{0.169} & $-$0.60 & $+$0.14 & 0.287 \\
qwen3:4b & 0.569 & 0.0\% & \underline{0.467} & 0.699 & 0.986 & \underline{0.081} & $-$0.57 & $+$0.34 & 0.274 \\
llama3.2:3b & 0.467 & 1.5\% & \underline{0.234} & 0.605 & 0.931 & \underline{0.275} & $-$0.51 & $+$0.14 & 0.254 \\
gemma4:e2b & 0.590 & 1.7\% & \underline{0.450} & 0.797 & 0.993 & \underline{0.016} & $-$0.66 & $+$0.32 & 0.233 \\
deepseek-r1:32b & 0.556 & 2.4\% & 0.953 & \underline{0.087} & \underline{0.349} & 0.942 & $+$0.73 & $+$0.14 & 0.218 \\
qwen3:1.7b & 0.517 & 1.5\% & 0.835 & \underline{0.286} & \underline{0.138} & 0.649 & $+$0.53 & $+$0.02 & 0.212 \\
gemma3:1b & 0.537 & 0.2\% & 0.967 & \underline{0.166} & \underline{0.101} & 0.790 & $+$0.74 & $+$0.06 & 0.133 \\
qwen3:0.6b & 0.548 & 0.0\% & 0.997 & \underline{0.160} & \underline{0.030} & 0.878 & $+$0.84 & $-$0.01 & 0.095 \\
deepseek-r1:14b & 0.477 & 10.6\% & 0.905 & \underline{0.014} & \underline{0.134} & 0.889 & $+$0.82 & $+$0.07 & 0.074 \\
deepseek-r1:7b & 0.202 & 61.5\% & \underline{0.415} & \underline{0.002} & \underline{0.000} & \underline{0.357} & $+$0.39 & $+$0.03 & 0.001 \\
\bottomrule
\end{tabular}
\end{table*}

 \begin{figure}[!htb]
\centering
\includegraphics[width=\linewidth]{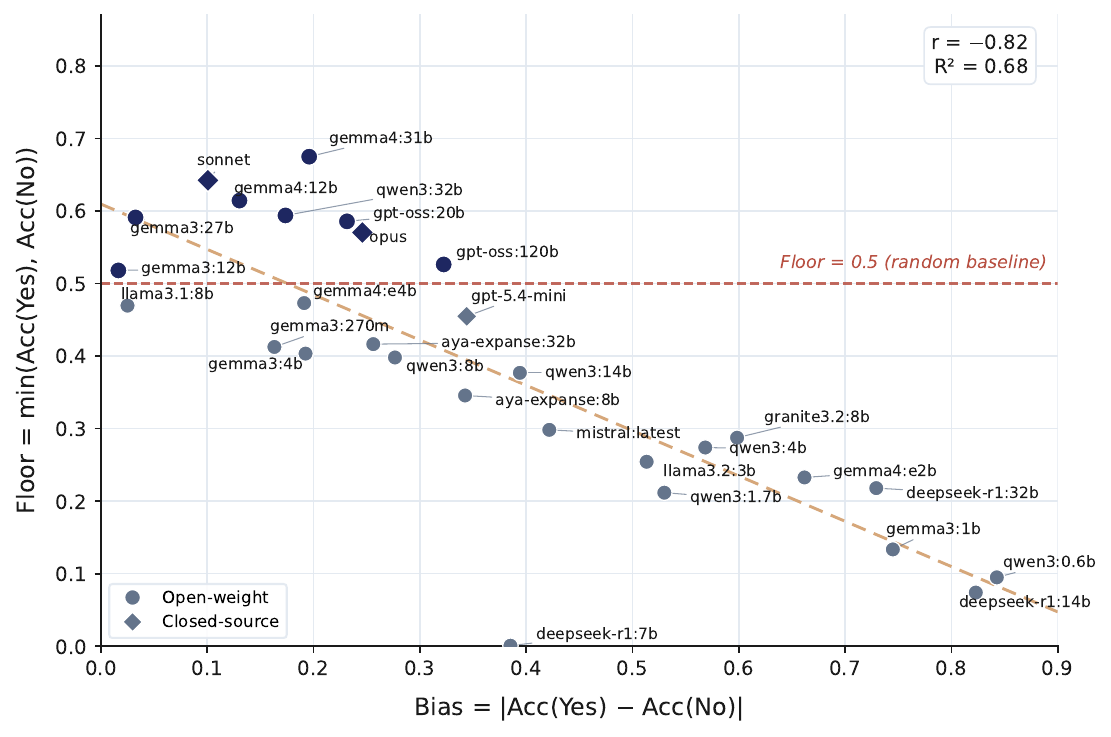}
\caption{Competence Floor against answer bias ($|\text{Bias}|$) for all 29 models. The two are strongly inversely correlated ($r = -0.82$, $R^2 = 0.68$): the more a model is biased toward one answer, the lower its floor. Only nine models clear the 0.5 random baseline (dashed), and the most biased models (e.g.\ Qwen3:0.6B) sit near the floor's zero.}
\label{fig:bias_floor}
\end{figure}

 
\paragraph{Few models clear the floor.}
Only 9 of 29 models exceed the random Floor baseline of 0.5: Gemma4:31B (0.675), Sonnet~4.6 (0.642), Gemma4:12B (0.614), Qwen3:32B (0.594), Gemma3:27B (0.591), GPT-oss:20 (0.586), Opus~4.7 (0.570), GPT-oss:120B (0.526), and Gemma3:12B (0.518).
\alvininline{For "latest," can get get a version number or a date of release?}
Even these nine clear it only modestly, by about 2 to 18 points above the 0.5 baseline.
Eleven of the 29 models fall below a floor of 0.30.
Qwen3:0.6B is the textbook constant Yes responder: near-ceiling where Yes is correct (\vaff{} 0.997, \ineg{} 0.878), near-zero where No is (\vneg{} 0.160, \iaff{} 0.030), with 54.8\% overall accuracy the average of these extremes.
 
\paragraph{Competence differences are not only due to negation}
It is instructive to contrast the competence floor in affirmative vs. negated questions. Recomputing the competence floor from the affirmative cells with $\min(\text{\vaff}, \text{\iaff})$ achieves this.
This affirmative-only floor correlates $r = 0.90$ with the full floor, still only yielding 10 of 29 models above chance, and still leaving the biased models with the worst measures of competence (Qwen3:0.6B at 0.030).
Importantly, the answer bias is visible in the affirmative validity contrast alone, so negation is one way to expose it, not necessarily its cause.


\section{Analysis}
\label{sec:analysis}
 
\subsection{Robustness to Surface Variation}
\label{sec:analysis:robustness}

\nayera{Maybe we should dig deeper here before claiming anything about content; these are averages and could hide per-model or per-template pockets.}
A genuine reasoner would return the same answer when only the surface of a prompt changes and the logic is held fixed. We examine several such surface variations: the question form and how its negation is written (prefix vs.\ the word \textit{not}), the activity scenario, and the name's nationality and gender.  The tables here show the models that move most; full per-model tables are in Appendix~\ref{app:full}. 


\subsubsection{Question Form}
\label{sec:analysis:framesens}
\alvininline{The main issue here is that saying ``Does it not follow that...'' etc makes it sound like the answer should be yes.}
Among affirmative questions, the five question forms land within about four points of each other (67.6--71.4\%), a much smaller spread than what we see with negation (29.2--58.4\%; Table~\ref{tab:qform}).  The collapse is confined to the negated wrapper of the \textsc{follow} form, \textit{Does it not follow that\,\dots?}, which falls to 29.2\%, below the 50\% chance level and nearly 20 points under the next-lowest wrapper (48.8\%). The other four negated wrappers cluster together: Varying adjectives with \textit{Is it false that\,\dots?} (58.4\%), \textit{Is it incorrect that\,\dots?} (56.0\%) and \textit{Is it invalid to conclude that\,\dots?} (48.8\%), together with the proposition-level \textit{Is it not probable that\,\dots?} (56.5\%), all sit between 49 and 58\%.
Measured within each form as the affirmative$-$negated drop (polarity sensitivity), the gap is smallest for the \textsc{truth} form (\textit{Is it true that} vs.\ \textit{Is it false that}, $+0.125$) and largest for the \textsc{follow} form (\textit{Does it follow that} vs.\ \textit{Does it not follow that}, $+0.384$); the \textsc{follow} form is the only one whose negated wrapper leaves the shared range.

The collapse is broad but not uniform: 21 of 29 models score at least 10 points lower on \textit{does it not follow} than on the other negated wrappers, and the strongest models that fall furthest (Gemma4:31B, Sonnet~4.6 and Opus~4.7 each drop 60 or more points), while the handful that do not show it are near-random constant responders already poor in all cases.
Tellingly, the \textsc{direct} form's negated wrapper \textit{Is it not probable that\,\dots?} also contains the word \textit{not} yet holds at 56.5\%, inside that range, so the difficulty is specific to writing the negation as \textit{not} inside the entailment wrapper (\textit{does it \emph{not} follow}), not the word \textit{not} itself.

\begin{table*}[!htb]
\caption{Per-question-form results (all English models pooled). \textbf{Aff}/\textbf{Neg} are affirmative/negated accuracy (micro-averaged over all templates and models, uncertain counts as incorrect); \textbf{PS} is polarity sensitivity (Aff${}-{}$Neg). Sorted by PS. Under negation only the \textsc{follow} form's wrapper, \textit{Does it not follow that\,\dots?}, collapses to near chance; the other four stay in a tight band, including \textsc{direct} (\textit{Is it not probable that\,\dots?}), which also uses the word \textit{not}.}
\label{tab:qform}
\small
\centering
\begin{tabular}{@{}lllccc@{}}
\toprule
\textbf{Form} & \textbf{Affirmative wrapper} & \textbf{Negated wrapper} & \textbf{Aff} & \textbf{Neg} & \textbf{PS} \\
\midrule
\textsc{truth}   & Is it true that \dots?              & Is it false that \dots?               & .710 & .584 & $+$.125 \\
\textsc{correct} & Is it correct that \dots?           & Is it incorrect that \dots?           & .699 & .560 & $+$.139 \\
\textsc{direct}  & Is it probable that \dots?          & Is it not probable that \dots?        & .714 & .565 & $+$.150 \\
\textsc{valid}   & Is it valid to conclude that \dots? & Is it invalid to conclude that \dots? & .676 & .488 & $+$.188 \\
\textsc{follow}  & Does it follow that \dots?          & Does it not follow that \dots?        & .677 & .292 & $+$.384 \\
\bottomrule
\end{tabular}
\end{table*}

Per model, the same divergence shows up as a wide swing across forms; the nine floor-clearing models have question-form ranges from about 0.19 to 0.50 (Table~\ref{tab:framesens}), each concentrated in the \textit{does it not follow} form.

\begin{table*}[!htb]
\caption{Question-form sensitivity (models that move most). Overall accuracy in each of the five question forms (fraction correct, micro-averaged over all templates and both the affirmative and negated conditions; uncertain counts as incorrect); \textbf{Range} is the best$-$worst spread. The remaining 23 models have Range $0.02$--$0.32$; the smallest belong to committed responders (Qwen3:1.7B $0.02$, DeepSeek-R1:7B $0.06$). Full per-model table in Table~\ref{tab:framesens_full}.}
\label{tab:framesens}
\footnotesize
\centering
\begin{tabular}{@{}lcccccc@{}}
\toprule
\textbf{Model} & \textsc{truth} & \textsc{correct} & \textsc{valid} & \textsc{follow} & \textsc{direct} & \textbf{Range} \\
\midrule
gemma4:31b & 0.85 & 0.87 & 1.00 & 0.50 & 0.88 & 0.50 \\
Sonnet 4.6 & 0.82 & 0.80 & 0.88 & 0.50 & 0.74 & 0.38 \\
gemma4:12b & 0.80 & 0.83 & 0.86 & 0.52 & 0.82 & 0.34 \\
gpt-oss:20 & 0.85 & 0.84 & 0.72 & 0.51 & 0.77 & 0.33 \\
Opus 4.7 & 0.83 & 0.82 & 0.84 & 0.51 & 0.78 & 0.33 \\
gpt-oss:120b & 0.85 & 0.79 & 0.63 & 0.53 & 0.82 & 0.32 \\
\bottomrule
\end{tabular}
\end{table*}

To test whether the negation strategy itself drives this collapse, we added \textit{not}-word versions of the \textsc{correct} and \textsc{valid} forms (\textit{Is it not correct that\,\dots?}, \textit{Is it not valid to conclude that\,\dots?}), creating matched pairs where the only difference is whether the negation is written as a prefix (\textit{incorrect}/\textit{invalid}) or with the word \textit{not}.

\subsubsection{Prefix Negation versus \textit{not}-Negation}
\label{sec:results:negstyle}

Since the \textit{not} negation is pragmatically ambiguous---it has a possible interpretation as pining for agreement with a rhetorical question---while prefix negation is unambiguous, these results provide information about the effect of this pragmatic ambiguity versus an unambiguous control.
Comparing these matched pairs (prefix \textit{incorrect}/\textit{invalid} against \textit{not correct}/\textit{not valid}), the prefix form yields 52.3\% accuracy versus 33.6\% for the \textit{not} form, an 18.7-point gap, suggesting that the ambiguous form confuses the model.
Table~\ref{tab:negstyle_full} presents per-model results.
Each model contributes 2,400 paired observations per negation style.
The per-model gaps are far larger: Sonnet drops 61.1 points, Opus 59.1, Qwen3:32B 40.3.
The direction is not uniform: 25 of 29 models perform better with the prefix form, but 4 (Qwen3:0.6B, Gemma3:1B, Qwen3:1.7B, Llama3.2:3B) show the reverse.

\begin{table*}[!htb]
\caption{Negation written as a prefix vs.\ with the word \textit{not}, for all 29 models. Prefix${}={}$\textit{incorrect}/\textit{invalid} (negation fused into the adjective); \textit{not}${}={}$\textit{not correct}/\textit{not valid} (a separate ``not''). Both sides are negated questions with the same proposition and gold answer; only the negation surface form differs. $\Delta={}$Prefix accuracy${}-{}$\textit{not} accuracy. The Yes-rate columns give the model's percentage of Yes answers under each form.}
\label{tab:negstyle_full}
\small
\centering
\begin{tabular}{lccccc}
\toprule
 & \multicolumn{3}{c}{\textbf{Accuracy}} & \multicolumn{2}{c}{\textbf{Yes-rate (\%)}} \\
\cmidrule(lr){2-4}\cmidrule(lr){5-6}
\textbf{Model} & \textbf{Prefix} & \textbf{\textit{not}} & \textbf{$\Delta$} & \textbf{Prefix} & \textbf{\textit{not}} \\
\midrule
gemma4:12b & 0.776 & 0.137 & $+$0.639 & 9.9 & 75.6 \\
Sonnet 4.6 & 0.785 & 0.175 & $+$0.611 & 23.5 & 84.5 \\
Opus 4.7 & 0.781 & 0.208 & $+$0.572 & 10.0 & 66.4 \\
gemma4:31b & 0.915 & 0.372 & $+$0.543 & 18.1 & 56.5 \\
qwen3:32b & 0.666 & 0.263 & $+$0.403 & 37.2 & 51.2 \\
gpt-5.4-mini & 0.632 & 0.298 & $+$0.334 & 23.8 & 58.3 \\
aya-expanse:32b & 0.503 & 0.193 & $+$0.310 & 37.4 & 87.5 \\
gpt-oss:20 & 0.723 & 0.426 & $+$0.297 & 32.3 & 36.0 \\
granite3.2:8b & 0.727 & 0.440 & $+$0.286 & 10.3 & 35.8 \\
gpt-oss:120b & 0.695 & 0.446 & $+$0.250 & 23.1 & 34.9 \\
gemma3:27b & 0.425 & 0.182 & $+$0.243 & 53.6 & 77.5 \\
gemma3:12b & 0.502 & 0.290 & $+$0.212 & 28.5 & 66.8 \\
llama3.1:8b & 0.510 & 0.356 & $+$0.154 & 40.5 & 55.5 \\
aya-expanse:8b & 0.441 & 0.300 & $+$0.140 & 64.0 & 83.4 \\
qwen3:4b & 0.531 & 0.417 & $+$0.113 & 26.4 & 35.9 \\
gemma4:e2b & 0.679 & 0.570 & $+$0.108 & 5.6 & 21.8 \\
qwen3:8b & 0.361 & 0.255 & $+$0.106 & 62.8 & 82.6 \\
qwen3:14b & 0.649 & 0.560 & $+$0.089 & 30.8 & 26.4 \\
gemma4:e4b & 0.458 & 0.421 & $+$0.036 & 38.0 & 66.1 \\
deepseek-r1:7b & 0.103 & 0.077 & $+$0.026 & 37.8 & 32.8 \\
gemma3:270m & 0.443 & 0.428 & $+$0.015 & 50.9 & 53.8 \\
deepseek-r1:14b & 0.250 & 0.236 & $+$0.014 & 93.7 & 91.1 \\
mistral:7b & 0.294 & 0.284 & $+$0.010 & 74.2 & 83.1 \\
gemma3:4b & 0.531 & 0.526 & $+$0.005 & 50.0 & 60.3 \\
deepseek-r1:32b & 0.261 & 0.259 & $+$0.002 & 98.3 & 98.3 \\
qwen3:0.6b & 0.351 & 0.359 & $-$0.008 & 82.7 & 88.2 \\
gemma3:1b & 0.245 & 0.257 & $-$0.012 & 96.4 & 93.5 \\
qwen3:1.7b & 0.386 & 0.425 & $-$0.039 & 79.0 & 69.5 \\
llama3.2:3b & 0.547 & 0.591 & $-$0.044 & 31.1 & 17.1 \\
\bottomrule
\end{tabular}
\end{table*}

\subsubsection{Activity}
\label{sec:analysis:activity}

Replacing the scenario verb phrases (``party'', ``event'', ``gathering'', ``meeting'', or ``celebration'') attenuates accuracy. The average per-model range across the five activities is 3.4 pts, relatively minor compared to negation and question-form effects. Only DeepSeek-R1:7b moves substantially (13.2 pts). As with nationality, this is a refusal pattern, its refusal rate ranging from 46\% to 72\% across the five activities while accuracy on the items it answers stays near 52\%; so the model is sensitive to the verb but expresses that as refusal to answer rather than as different reasoning~(Table~\ref{tab:activity}).

\begin{table*}[!htb]
\caption{Accuracy by activity scenario (models that move most; uncertain counts as incorrect). \textbf{Range} is the spread across the five activities. The remaining 24 models have Range ${}\le0.04$ (mean Range $=0.034$). Full per-model table in Table~\ref{tab:activity_full}.}
\label{tab:activity}
\footnotesize
\centering
\begin{tabular}{@{}lcccccc@{}}
\toprule
\textbf{Model} & \textbf{Party} & \textbf{Event} & \textbf{Gather} & \textbf{Meeting} & \textbf{Celebr.} & \textbf{Range} \\
\midrule
deepseek-r1:7b & 0.25 & 0.28 & 0.17 & 0.17 & 0.15 & 0.13 \\
qwen3:8b & 0.60 & 0.64 & 0.58 & 0.61 & 0.57 & 0.06 \\
llama3.1:8b & 0.43 & 0.49 & 0.49 & 0.44 & 0.46 & 0.06 \\
gpt-oss:20 & 0.77 & 0.73 & 0.72 & 0.75 & 0.71 & 0.06 \\
gemma3:27b & 0.65 & 0.67 & 0.63 & 0.66 & 0.62 & 0.05 \\
\bottomrule
\end{tabular}
\end{table*}

\subsubsection{Nationality}
\label{sec:analysis:nationality}

We find biases are introduced by varying gender, nationality, and activity with every model.  As a control, we also use variables $X$ and $Y$ instead of people's names to compare against.
Names are drawn from seven nationality groups, and Table~\ref{tab:nationality} reports each nationality's accuracy \emph{relative to} the abstract variable ($X,Y$) baseline. (A positive value means the model does better on that nationality than on the variables.) For 23 of 29 the largest per-nationality gap is at most 4.7 points. The clear exception is DeepSeek-R1:7b, which sits 13 to 34 points below variables on every nationality, being worst on Japanese. It frequently refuses to answer the question, but on the items it actually answers, its accuracy across nationalities spans about 6 points (51--57\%), against roughly 21 points on the headline accuracy; the gap comes instead from how often it refuses, and that refusal rate itself swings sharply with nationality, from 46\% on Indian names to 87\% on Japanese. 

\begin{table*}[!htb]
\caption{Accuracy on each nationality group \emph{minus} accuracy on the abstract letter-variable ($X,Y$) baseline, in percentage points (models that move most; uncertain counts as incorrect). Positive${}={}$higher on that nationality than on XY. Ind=Indian, Rus=Russian, Jpn=Japanese, Afr=African, Ger=German, Fre=French, Amr=American. $\max|\Delta|$ is the largest absolute gap. The remaining 23 models have $\max|\Delta|\le4.7$ points. Full per-model table in Table~\ref{tab:nationality_full}.}
\label{tab:nationality}
\footnotesize
\centering
\begin{tabular}{@{}lcccccccc@{}}
\toprule
\textbf{Model} & \textbf{Ind} & \textbf{Rus} & \textbf{Jpn} & \textbf{Afr} & \textbf{Ger} & \textbf{Fre} & \textbf{Amr} & \textbf{max}\,$|\Delta|$ \\
\midrule
deepseek-r1:7b & $-$13.5 & $-$23.7 & $-$34.1 & $-$20.2 & $-$26.3 & $-$27.8 & $-$23.4 & 34.1 \\
deepseek-r1:32b & $-$1.9 & $-$0.5 & $-$1.6 & $-$2.2 & $-$0.8 & $-$9.5 & $-$1.7 & 9.5 \\
llama3.2:3b & $-$3.6 & $-$2.2 & $-$9.1 & $-$2.7 & $-$2.8 & $-$8.2 & $-$4.0 & 9.1 \\
llama3.1:8b & $-$2.2 & $+$1.9 & $+$3.2 & $+$7.5 & $+$3.8 & $+$4.8 & $+$6.2 & 7.5 \\
gemma3:12b & $-$2.1 & $-$3.6 & $-$3.9 & $-$3.8 & $-$2.2 & $-$4.8 & $-$2.8 & 4.8 \\
Opus 4.7 & $+$2.1 & $+$4.8 & $+$0.9 & $+$1.1 & $+$1.3 & $+$0.0 & $+$1.1 & 4.8 \\
\bottomrule
\end{tabular}
\end{table*}

\subsubsection{Gender}
\label{sec:analysis:gender}

Names are balanced by gender, and Table~\ref{tab:gender} reports each gender's accuracy relative to the variable ($X,Y$) control. Women's and men's names score within about 3 points of variables for every model except DeepSeek-R1:7b, which 
exhibits the same refusal behavior seen for nationality. The man-woman name gap itself average 0.6 percentage points; the largest being DeepSeek-R1:32b ($+2.3$ toward men's names) and GPT-5.4-mini ($+2.2$ toward women's names).

\begin{table}[t]
\caption{Accuracy on female and male names \emph{minus} accuracy on the variable ($X,Y$) control, in percentage points (models that move most; uncertain counts as incorrect). Positive${}={}$higher than $X,Y$; the male$-$female gap is the difference between the two columns. $\max|\Delta|$ is the larger absolute gap. The remaining 23 models have $\max|\Delta|\le3.0$ points. Full per-model table in Table~\ref{tab:gender_full}.}
\label{tab:gender}
\footnotesize
\centering
\begin{tabular}{@{}lccc@{}}
\toprule
\textbf{Model} & \textbf{Female} & \textbf{Male} & \textbf{max}\,$|\Delta|$ \\
\midrule
deepseek-r1:7b & $-$24.3 & $-$24.0 & 24.3 \\
llama3.2:3b & $-$4.8 & $-$4.5 & 4.8 \\
llama3.1:8b & $+$4.0 & $+$3.2 & 4.0 \\
deepseek-r1:32b & $-$3.8 & $-$1.4 & 3.8 \\
gemma3:12b & $-$3.6 & $-$3.0 & 3.6 \\
qwen3:14b & $-$2.7 & $-$3.1 & 3.1 \\
\bottomrule
\end{tabular}
\end{table}

\subsection{Conjunction Variants}
\label{sec:analysis:conjvariants}
 
We test two conjunction inferences, each written two ways that keep the logic fixed (Table~\ref{tab:conjexamples}), with per-model accuracies in Table~\ref{tab:conjdiffs}. 
Recall that distribution over conjunction (DOC) is valid while conjunctivitis is invalid~(\S\ref{sec:benchmark:templates}).
By DOC, $Pr(A\land B)\implies Pr(A)\land Pr(B)$.  To test for consistent reasoning, we vary whether we ask the model about the truth of $Pr(A)$ or $P(B)$, since logically both are true.

On DOC examples, affirmative questions differ by about 5 points on average, and 25 of 29 models stay within 10 points, a large gap for such a simple variation.
For conjunctivitis examples how the two premises are worded matters even more. We see this by testing two conditions: one with two people engaging in the same activity and one with two people engaging in different activities.  On affirmative items, where the model has to reject the inference, the same-activity wording (two people at one activity) is easier to reject than the dual-activity wording (one person at two activities) by a mean of 12.8 points (up to 96 points for Qwen3:14B, and Sonnet drops from 96.5\% to 56.5\%). This gap resides entirely in the reject condition. On negated items, where the answer is \textit{Yes}, the two wordings yield very close results (0.3 points), so pooling the two conditions dilutes the effect to 6.6 points, which is why we report it on the affirmative items.

\begin{table*}[t]
\caption{The two conjunction inference types, each shown in two surface realizations (affirmative condition; the negated condition swaps \textit{Is it true that} for \textit{Is it false that} and flips the gold answer). The distribution-over-conjunction rows differ only in which name is queried; the conjunctivitis rows differ only in whether the two premises describe two people at one activity or one person at two activities.}
\label{tab:conjexamples}
\footnotesize
\centering
\begin{tabular}{@{}p{2.7cm}p{8cm}c@{}}
\toprule
\textbf{Template (logic)} & \textbf{Example (affirmative condition)} & \textbf{Gold} \\
\midrule
Dist.\ over conjunction, valid, query first name & It is probable that Savir and Ashwin will be at the party. \textit{Is it true that it is probable that Savir will be at the party?} & Yes \\
\addlinespace
Dist.\ over conjunction, valid, query second name & It is probable that Savir and Ashwin will be at the party. \textit{Is it true that it is probable that Ashwin will be at the party?} & Yes \\
\addlinespace
Conjunctivitis, invalid, same activity & It is probable that Savir will be at the party. It is probable that Ashwin will be at the party. \textit{Is it true that it is probable that Savir and Ashwin will be at the party?} & No \\
\addlinespace
Conjunctivitis, invalid, dual activity & It is probable that Savir will attend the event. It is probable that Savir will give a presentation. \textit{Is it true that it is probable that Savir will attend the event and give a presentation?} & No \\
\bottomrule
\end{tabular}
\end{table*}

\begin{table*}[!htb]
\caption{Per-model accuracy on the two conjunction templates under a wording change that leaves the logic fixed (affirmative items; uncertain counts as incorrect). \textbf{Conjunctivitis} is invalid (correct answer No), so its accuracy is the rejection rate; \textbf{Same} states the two premises as two people at one activity and \textbf{Dual} as one person at two activities. \textbf{Distribution over conjunction} is valid (correct answer Yes); \textbf{Name~1} and \textbf{Name~2} ask about the first or second name in the shared premise. $\Delta$ is the accuracy difference between the two wordings. Rewriting the conjunctivitis premises moves accuracy by a mean of 12.8 points (up to 96), while which name distribution over conjunction asks about moves it by only 4.9 on average. Sorted by the conjunctivitis $\Delta$.}
\label{tab:conjdiffs}
\footnotesize
\centering
\begin{tabular}{@{}lccc@{\hspace{1.4em}}ccc@{}}
\toprule
& \multicolumn{3}{c}{\textbf{Conjunctivitis (reject rate)}} & \multicolumn{3}{c}{\textbf{Dist.\ over conjunction}} \\
\cmidrule(lr){2-4}\cmidrule(lr){5-7}
\textbf{Model} & \textbf{Same} & \textbf{Dual} & \textbf{$\Delta$} & \textbf{Name~1} & \textbf{Name~2} & \textbf{$\Delta$} \\
\midrule
qwen3:14b & 1.00 & 0.04 & $+$0.96 & 0.90 & 0.98 & $-$0.09 \\
qwen3:8b & 0.51 & 0.00 & $+$0.51 & 1.00 & 1.00 & $+$0.00 \\
Sonnet 4.6 & 0.96 & 0.56 & $+$0.40 & 1.00 & 1.00 & $+$0.00 \\
gemma3:12b & 0.35 & 0.01 & $+$0.35 & 1.00 & 1.00 & $+$0.00 \\
gemma3:27b & 0.70 & 0.35 & $+$0.35 & 1.00 & 1.00 & $+$0.00 \\
gemma4:12b & 0.88 & 0.55 & $+$0.33 & 0.97 & 1.00 & $-$0.03 \\
granite3.2:8b & 0.96 & 0.71 & $+$0.26 & 0.68 & 0.82 & $-$0.14 \\
qwen3:32b & 0.83 & 0.59 & $+$0.24 & 0.97 & 0.99 & $-$0.02 \\
deepseek-r1:32b & 0.27 & 0.04 & $+$0.23 & 1.00 & 1.00 & $-$0.00 \\
gpt-oss:20 & 0.90 & 0.69 & $+$0.21 & 0.86 & 0.84 & $+$0.02 \\
gpt-oss:120b & 0.89 & 0.71 & $+$0.18 & 0.95 & 0.93 & $+$0.02 \\
gemma4:e4b & 0.49 & 0.40 & $+$0.09 & 0.94 & 0.98 & $-$0.04 \\
gemma3:4b & 0.09 & 0.01 & $+$0.09 & 0.79 & 0.95 & $-$0.17 \\
deepseek-r1:14b & 0.08 & 0.00 & $+$0.08 & 0.99 & 0.98 & $+$0.01 \\
llama3.1:8b & 0.13 & 0.05 & $+$0.08 & 0.55 & 0.64 & $-$0.08 \\
aya-expanse:32b & 0.02 & 0.00 & $+$0.02 & 1.00 & 1.00 & $+$0.00 \\
qwen3:1.7b & 0.02 & 0.00 & $+$0.02 & 1.00 & 0.99 & $+$0.01 \\
gpt-5.4-mini & 0.95 & 0.94 & $+$0.01 & 0.23 & 0.66 & $-$0.43 \\
aya-expanse:8b & 0.00 & 0.00 & $+$0.00 & 1.00 & 1.00 & $+$0.00 \\
deepseek-r1:7b & 0.00 & 0.00 & $+$0.00 & 0.23 & 0.19 & $+$0.04 \\
gemma3:1b & 0.00 & 0.00 & $+$0.00 & 1.00 & 1.00 & $+$0.00 \\
gemma4:31b & 1.00 & 1.00 & $+$0.00 & 1.00 & 1.00 & $+$0.00 \\
qwen3:0.6b & 0.00 & 0.00 & $+$0.00 & 1.00 & 0.99 & $+$0.01 \\
qwen3:4b & 1.00 & 1.00 & $+$0.00 & 0.99 & 0.97 & $+$0.02 \\
Opus 4.7 & 0.99 & 1.00 & $-$0.01 & 0.82 & 0.95 & $-$0.14 \\
gemma4:e2b & 0.98 & 0.99 & $-$0.01 & 0.92 & 0.97 & $-$0.05 \\
llama3.2:3b & 0.86 & 0.89 & $-$0.02 & 0.43 & 0.48 & $-$0.05 \\
mistral:7b & 0.00 & 0.06 & $-$0.06 & 0.99 & 1.00 & $-$0.01 \\
gemma3:270m & 0.07 & 0.67 & $-$0.60 & 0.49 & 0.47 & $+$0.02 \\
\midrule
\textbf{Mean} & 0.52 & 0.39 & $+$0.13 & 0.85 & 0.89 & $-$0.04 \\
\bottomrule
\end{tabular}
\end{table*}

\subsection{Per-Template Breakdown}
\label{sec:analysis:templates}
 
The 13 templates greatly vary in accuracy (42.8\% for Conjunctivitis to 74.3\% for Probably to Might).
The key finding is a dissociation (Table~\ref{tab:template_analysis}): the answer bias varies less across templates than polarity sensitivity does ($|\text{Bias}|$ std $= 0.056$ versus PS std $= 0.126$, about twice as much), and it stays high on every template (0.42 to 0.62).
The bias is thus a model-level property carried across all templates, while negation sensitivity is template-dependent.
Model biases also change direction based on template: on the easiest template (Probably to Might) 28 of 29 models are biased toward \textit{Yes}, so nearly all give the same answer. On the two hardest templates the panel splits roughly in half (Conjunctivitis 13 yes-biased vs.\ 16 no-biased; Might to Probably 12 vs.\ 17): the template exerts no shared pull, so which answer a model gives is set mostly by its own bias rather than by the content of the inference.
 
\begin{table*}[!htb]
\caption{Per-template analysis (all English models pooled), grouped by validity. \textbf{Acc} is overall accuracy on the template, a micro-average with uncertain counted as incorrect. $|\textbf{Bias}|$ is the mean absolute answer bias (from the yes-rate), \textbf{PS} the mean polarity sensitivity, and \textbf{Agree.} the fraction of models sharing the majority bias direction.}
\label{tab:template_analysis}
\small
\centering
\begin{tabular}{llcccc}
\toprule
\textbf{Type} & \textbf{Template} & \textbf{Acc} & $|\textbf{Bias}|$ & \textbf{PS} & \textbf{Agree.} \\
\midrule
Valid & Chancy Disjunction Introduction & .574 & .539 & $+$.098 & 62.1\% \\
 & Chancy Modus Ponens & .695 & .453 & $+$.276 & 72.4\% \\
 & Chancy Modus Tollens & .547 & .526 & $+$.095 & 58.6\% \\
 & Complement Transfer & .604 & .496 & $+$.183 & 65.5\% \\
 & Conditional to Comparative & .540 & .497 & $+$.134 & 62.1\% \\
 & Distribution over Conjunction & .695 & .418 & $+$.351 & 89.7\% \\
 & Must to Probably & .633 & .509 & $+$.172 & 62.1\% \\
 & Positive Form Transfer & .674 & .488 & $+$.023 & 65.5\% \\
 & Probably to Might & .743 & .463 & $+$.417 & 96.6\% \\
 & Probably to not probably not & .578 & .548 & $+$.335 & 79.3\% \\
\midrule
Invalid & Conjunctivitis & .428 & .619 & $+$.047 & 55.2\% \\
 & Might to Probably & .447 & .608 & $+$.097 & 58.6\% \\
 & Probably to Certain & .683 & .462 & $+$.342 & 79.3\% \\
\bottomrule
\end{tabular}
\end{table*}
 

\subsection{Model Scale}
\label{sec:analysis:scaling}
 
Four families provide size variants (Table~\ref{tab:scaling}).
The answer bias tends to fall as models grow, most clearly in Qwen~3 (Spearman $\rho = -0.89$ between $|$Bias$|$ and size) and more weakly in Gemma~3 ($\rho = -0.60$) and Gemma~4 ($\rho = -0.40$), but no family is monotonic (Gemma~3, for instance, spikes at 1B before falling again). DeepSeek-R1 runs the other way, its bias rising with size ($\rho = +0.50$). Thus scaling lowers the bias on average in three of four families, never cleanly, and not at all in the fourth.
The \textsc{follow} question form (the word \textit{not}) shows little scaling improvement across families, suggesting the difficulty with \textit{not} is not resolved by additional parameters.
 
\begin{table}[!htb]
\caption{Scaling within size-graded model families (English). \textbf{Acc} is overall accuracy with uncertain responses counted as incorrect. $|\textbf{Bias}|$ is the magnitude of the signed answer bias and \textbf{PS} the polarity sensitivity, both defined in \S\ref{sec:benchmark:metrics} (Bias${}=\text{Acc(Yes)}-\text{Acc(No)}$; PS${}=\text{Acc(affirmative)}-\text{Acc(negated)}$, so positive${}={}$better on affirmative). \textbf{Unc\%} is the uncertain rate. Sizes are listed smallest to largest within each family.}
\label{tab:scaling}
\small
\centering
\begin{tabular}{llcccc}
\toprule
\textbf{Family} & \textbf{Size} & \textbf{Acc} & $|\textbf{Bias}|$ & \textbf{PS} & \textbf{Unc\%} \\
\midrule
Gemma~3 & 270M & 0.500 & 0.163 & +0.107 & 0.0 \\
 & 1B & 0.537 & 0.745 & +0.056 & 0.2 \\
 & 4B & 0.529 & 0.192 & -0.113 & 0.0 \\
 & 12B & 0.615 & 0.016 & +0.209 & 0.0 \\
 & 27B & 0.648 & 0.033 & +0.411 & 1.0 \\
\midrule
Gemma~4 & E2B & 0.590 & 0.662 & +0.315 & 1.7 \\
 & E4B & 0.608 & 0.191 & +0.189 & 0.0 \\
 & 12B & 0.764 & 0.130 & +0.343 & 0.0 \\
 & 31B & 0.818 & 0.196 & +0.412 & 0.0 \\
\midrule
Qwen~3 & 0.6B & 0.548 & 0.843 & -0.005 & 0.0 \\
 & 1.7B & 0.517 & 0.530 & +0.019 & 1.5 \\
 & 4B & 0.569 & 0.568 & +0.337 & 0.0 \\
 & 8B & 0.599 & 0.277 & +0.238 & 0.0 \\
 & 14B & 0.605 & 0.394 & +0.053 & 0.0 \\
 & 32B & 0.718 & 0.174 & +0.279 & 0.0 \\
\midrule
DeepSeek-R1 & 7B & 0.202 & 0.385 & +0.028 & 61.5 \\
 & 14B & 0.477 & 0.823 & +0.068 & 10.6 \\
 & 32B & 0.556 & 0.729 & +0.137 & 2.4 \\
\bottomrule
\end{tabular}
\end{table}

\begin{figure*}[!htb]
\centering
\includegraphics[width=\linewidth]{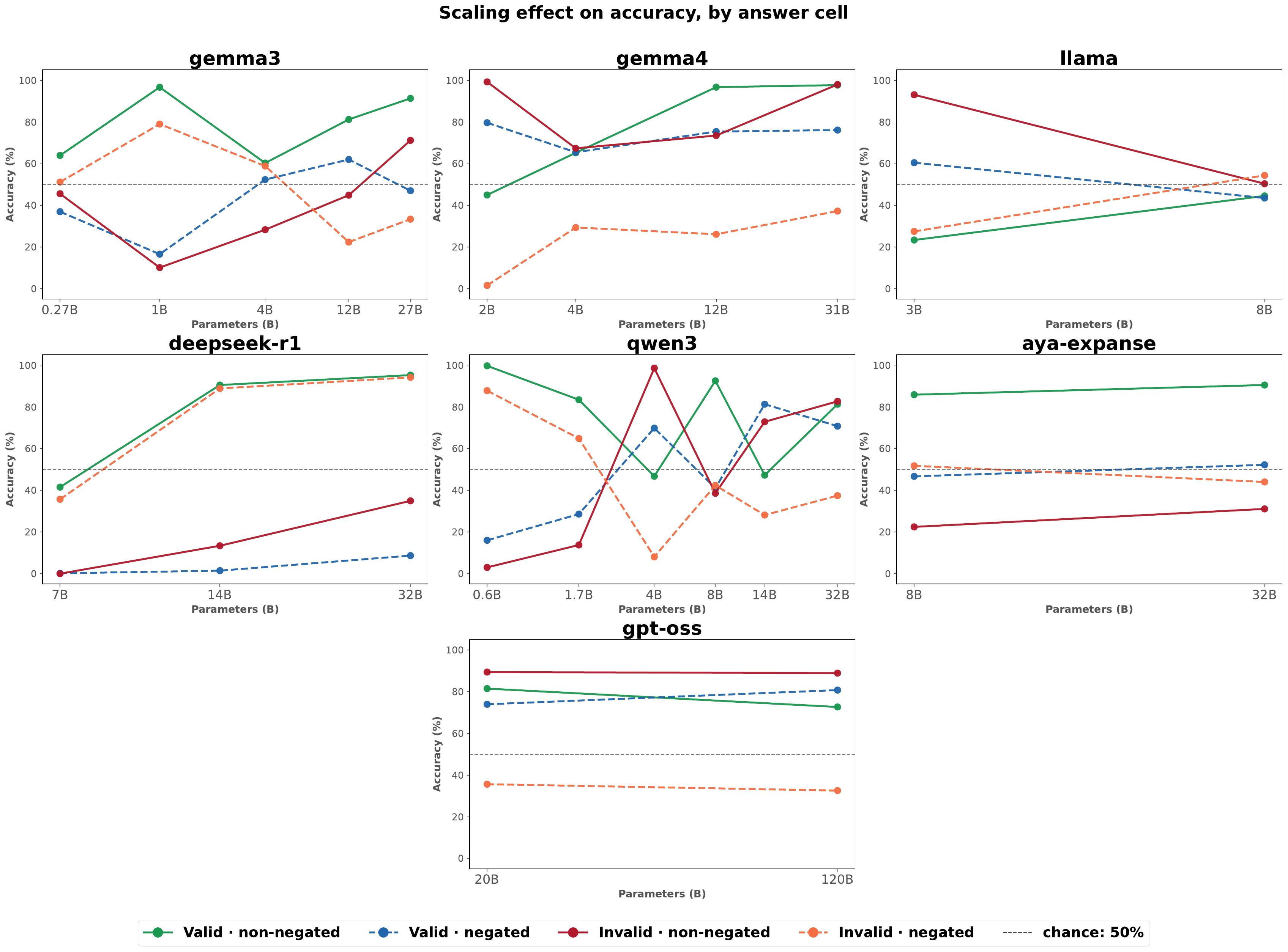}
\caption{Effect of scaling on accuracy by answer cell (valid/invalid $\times$ negated/non-negated) across model families. Dashed line marks 50\% chance.  Scaling does not consistently improve accuracy.\textcolor{red}}
\label{fig:scaling}
\end{figure*}


\section{Discussion}

Overall, we find widespread biases across multiple dimensions, including name origin, gender, and various perturbations to question formation.  More generally, we find definitive evidence that while some models manage to perform above chance, many do not even reach this low bar, and LLMs do not, in general, make basic EM inferences consistently, making them inappropriate for use in question-answering settings that require synthesizing epistemic statements to make valid conclusions.

\label{sec:discussion}
\paragraph{What accuracy can hide}
Most models show a fixed answer bias for \textit{Yes} or \textit{No} belied by aggregate accuracy.
(Valid) reasoning requires both accepting valid inferences and rejecting invalid ones, but we demonstrate conclusively that state of the art models fail to do this.  We suggest that future benchmarks using binary questions to evaluate reasoning also report the competence floor, which is cheap to compute and is not inflated by an answer bias.  While precision and recall also attempt to address this, they are more difficult to interpret in this specific scenario.
 

\paragraph{Negation is not the main source of answer bias}
Two of our probes use negation, and answer biases do not depend on it: the affirmative-only competence floor correlates $r = 0.90$ with the full (affirmative+negation) floor, and the most biased models (Qwen3:0.6B at 0.030) score the worst at both. 
\paragraph{What the floor does and does not say}
The floor measures whether a model performs above chance on both answer classes.
The invalid templates are uncontroversial: the conjunction fallacy, possibility not entailing probability, and probability not entailing certainty hold under any reasonable semantics.  However, some valid templates rest on stronger commitments. Conditional-to-comparative and probably-to-not-probably-not depend on specific axioms a competent reasoner could contest, though this is not responsible for the results: recomputing the floor with those two templates leaves the number of models with accuracy above 0.5 unchanged at 9 of 29 and moves every above-floor model by at most 0.03 (Sonnet 0.642 to 0.665, Opus 0.570 to 0.567, Qwen3:32B 0.594 to 0.606). Restricting further to a conservative core of only uncontestable inferences (the three invalid templates plus the four most basic valid ones, where \textit{probably}~$\phi$ entails \textit{might}~$\phi$, \textit{must}~$\phi$ entails \textit{probably}~$\phi$, distribution over conjunction, and chancy disjunction introduction) tells the same story (10 of 29 above 0.5), and in both restrictions the most biased models remain at the bottom (DeepSeek-R1:7B at 0.00).
Biased responders fail the uncontroversial, invalid templates regardless of semantics, so their low floor cannot be attributed to legitimately contested axioms.
An expert or human baseline calibrating how much residual sub-floor behavior is theory disagreement is the natural next step.
 
\paragraph{Limitations}
Our evaluation is zero-shot and English-only, and parses the first answer token rather than a free-form justification; whether these patterns persist under longer generation or in other languages is left to future work.

\section{Conclusion}
\label{sec:conclusion}
 
We have introduced a benchmark for inference over probability operators that separates a model's answer bias from proposition-tracking, demonstrating conclusively that models consistently fail at zero-shot reasoning under trivial prompts with an unambiguous correct response.
Most models show a fixed bias toward \textit{Yes} or \textit{No}: only 9 of 29 models exceed a competence floor with an absolute random baseline, and the result reproduces from the affirmative validity contrast alone, so it does not appear to depend on negation. Negation variation exposes a strong bias, where prefix and \textit{not} phrasing of the same content differ by up to 64 points.
Taken together, these results strongly suggest that part of what is commonly reported as ``reasoning'' ability may reflect answer bias and is not the result of a coherent internal inference model. A competence floor with an absolute baseline is a useful and easy-to-calculate complement to accuracy as a summary of logical competence. 
    Our prompts are simple and formulaic, and all information necessary to answer correctly is provided in the prompts, suggesting that more subtle ways of expressing the same premises would not lead to successful inferences. Since most inference work limited to English, future work can extend this to other languages.




 
\bibliographystyle{compling}
\bibliography{references}
 
\appendix
 
\section{Additional Results}
\label{app:full}
 

\begin{table*}[!htb]
\caption{Overall accuracy under five pooling schemes (frames \textit{not correct}/\textit{not valid}/conclude excluded, templates folded to 13, uncertain counts as incorrect). All five score the same responses and differ only in weighting. \textbf{micro} is the headline flat mean over every response. \textbf{prompt-bal}, \textbf{tmpl-bal} and \textbf{frame-bal} weight each prompt, template or question form equally. \textbf{cat-avg} weights the four validity$\times$negation cells equally, $(\text{V.Aff}+\text{V.Neg}+\text{I.Aff}+\text{I.Neg})/4$. Frame-balancing equals micro by construction and prompt-balancing is within a fraction of a point of it; only cat-avg moves the number appreciably, always toward 0.5, because it stops the abundant easy valid cells from dominating the rare invalid-negation cell. \textbf{Spread} is the max$-$min across the five schemes for each model. The mean weighting spread is 4.7 points, and averaged over models cat-avg sits 3.2 points below micro. Sorted by micro.}
\label{tab:pooling}
\footnotesize
\centering
\begin{tabular}{@{}lcccccc@{}}
\toprule
\textbf{Model} & \textbf{micro} & \textbf{prompt-bal} & \textbf{tmpl-bal} & \textbf{frame-bal} & \textbf{cat-avg} & \textbf{Spread} \\
\midrule
gemma4:31b & 0.818 & 0.818 & 0.829 & 0.819 & 0.773 & 0.056 \\
gemma4:12b & 0.764 & 0.764 & 0.784 & 0.764 & 0.679 & 0.105 \\
Opus 4.7 & 0.756 & 0.782 & 0.769 & 0.756 & 0.693 & 0.089 \\
Sonnet 4.6 & 0.751 & 0.773 & 0.761 & 0.750 & 0.692 & 0.081 \\
gpt-oss:20 & 0.737 & 0.737 & 0.749 & 0.737 & 0.701 & 0.048 \\
gpt-oss:120b & 0.724 & 0.724 & 0.733 & 0.725 & 0.687 & 0.046 \\
qwen3:32b & 0.718 & 0.718 & 0.726 & 0.718 & 0.681 & 0.045 \\
gpt-5.4-mini & 0.656 & 0.656 & 0.668 & 0.656 & 0.627 & 0.041 \\
gemma3:27b & 0.648 & 0.648 & 0.665 & 0.648 & 0.607 & 0.058 \\
aya-expanse:32b & 0.624 & 0.624 & 0.652 & 0.625 & 0.545 & 0.107 \\
gemma3:12b & 0.615 & 0.615 & 0.631 & 0.615 & 0.526 & 0.105 \\
gemma4:e4b & 0.608 & 0.608 & 0.613 & 0.609 & 0.569 & 0.044 \\
granite3.2:8b & 0.607 & 0.607 & 0.606 & 0.608 & 0.587 & 0.021 \\
qwen3:14b & 0.605 & 0.605 & 0.604 & 0.605 & 0.574 & 0.031 \\
qwen3:8b & 0.599 & 0.599 & 0.607 & 0.600 & 0.536 & 0.071 \\
gemma4:e2b & 0.590 & 0.590 & 0.581 & 0.590 & 0.564 & 0.026 \\
aya-expanse:8b & 0.586 & 0.586 & 0.596 & 0.586 & 0.517 & 0.079 \\
qwen3:4b & 0.569 & 0.569 & 0.566 & 0.569 & 0.558 & 0.011 \\
deepseek-r1:32b & 0.556 & 0.556 & 0.559 & 0.557 & 0.583 & 0.027 \\
qwen3:0.6b & 0.548 & 0.548 & 0.552 & 0.548 & 0.516 & 0.036 \\
gemma3:1b & 0.537 & 0.537 & 0.535 & 0.537 & 0.506 & 0.031 \\
mistral:7b & 0.535 & 0.535 & 0.534 & 0.536 & 0.509 & 0.027 \\
gemma3:4b & 0.529 & 0.529 & 0.532 & 0.530 & 0.500 & 0.032 \\
qwen3:1.7b & 0.517 & 0.517 & 0.521 & 0.517 & 0.477 & 0.044 \\
gemma3:270m & 0.500 & 0.500 & 0.499 & 0.500 & 0.494 & 0.006 \\
deepseek-r1:14b & 0.477 & 0.477 & 0.472 & 0.476 & 0.486 & 0.014 \\
llama3.2:3b & 0.467 & 0.467 & 0.468 & 0.467 & 0.511 & 0.044 \\
llama3.1:8b & 0.463 & 0.463 & 0.481 & 0.464 & 0.482 & 0.019 \\
deepseek-r1:7b & 0.202 & 0.202 & 0.201 & 0.202 & 0.193 & 0.009 \\
\midrule
\textbf{Mean} & 0.597 & 0.598 & 0.603 & 0.597 & 0.565 & 0.047 \\
\bottomrule
\end{tabular}
\end{table*}

\begin{table*}[!htb]
\caption{Per-model question-form sensitivity, full 29-model table (uncertain counts as incorrect). Overall accuracy in each of the five question forms; \textbf{Range} is the best$-$worst spread. Ordered by competence floor; the faded rule separates the nine models that clear the 0.5 floor from those that do not (\S\ref{sec:analysis:framesens}).}
\label{tab:framesens_full}
\footnotesize
\centering
\begin{tabular}{@{}lcccccc@{}}
\toprule
\textbf{Model} & \textsc{truth} & \textsc{correct} & \textsc{valid} & \textsc{follow} & \textsc{direct} & \textbf{Range} \\
\midrule
gemma4:31b & 0.85 & 0.87 & 1.00 & 0.50 & 0.88 & 0.50 \\
Sonnet 4.6 & 0.82 & 0.80 & 0.88 & 0.50 & 0.74 & 0.38 \\
gemma4:12b & 0.80 & 0.83 & 0.86 & 0.52 & 0.82 & 0.34 \\
qwen3:32b & 0.83 & 0.72 & 0.72 & 0.52 & 0.80 & 0.31 \\
gemma3:27b & 0.73 & 0.62 & 0.67 & 0.53 & 0.69 & 0.19 \\
gpt-oss:20 & 0.85 & 0.84 & 0.72 & 0.51 & 0.77 & 0.33 \\
Opus 4.7 & 0.83 & 0.82 & 0.84 & 0.51 & 0.78 & 0.33 \\
gpt-oss:120b & 0.85 & 0.79 & 0.63 & 0.53 & 0.82 & 0.32 \\
gemma3:12b & 0.73 & 0.66 & 0.56 & 0.51 & 0.62 & 0.21 \\
\arrayrulecolor{black!30}\midrule\arrayrulecolor{black}
gemma4:e4b & 0.71 & 0.68 & 0.45 & 0.51 & 0.70 & 0.26 \\
llama3.1:8b & 0.50 & 0.58 & 0.35 & 0.28 & 0.61 & 0.32 \\
gpt-5.4-mini & 0.76 & 0.73 & 0.64 & 0.48 & 0.67 & 0.28 \\
aya-expanse:32b & 0.73 & 0.69 & 0.54 & 0.46 & 0.71 & 0.28 \\
gemma3:270m & 0.50 & 0.55 & 0.47 & 0.47 & 0.51 & 0.08 \\
gemma3:4b & 0.57 & 0.54 & 0.46 & 0.44 & 0.64 & 0.20 \\
qwen3:8b & 0.63 & 0.62 & 0.54 & 0.49 & 0.72 & 0.24 \\
qwen3:14b & 0.65 & 0.64 & 0.51 & 0.54 & 0.68 & 0.17 \\
aya-expanse:8b & 0.67 & 0.62 & 0.53 & 0.47 & 0.64 & 0.20 \\
mistral:7b & 0.61 & 0.52 & 0.48 & 0.43 & 0.63 & 0.20 \\
granite3.2:8b & 0.68 & 0.62 & 0.62 & 0.45 & 0.66 & 0.23 \\
qwen3:4b & 0.63 & 0.60 & 0.47 & 0.53 & 0.61 & 0.16 \\
llama3.2:3b & 0.49 & 0.51 & 0.42 & 0.45 & 0.46 & 0.09 \\
gemma4:e2b & 0.65 & 0.65 & 0.62 & 0.47 & 0.56 & 0.18 \\
deepseek-r1:32b & 0.56 & 0.52 & 0.55 & 0.56 & 0.60 & 0.07 \\
qwen3:1.7b & 0.50 & 0.52 & 0.53 & 0.52 & 0.51 & 0.02 \\
gemma3:1b & 0.55 & 0.51 & 0.50 & 0.50 & 0.62 & 0.12 \\
qwen3:0.6b & 0.50 & 0.51 & 0.59 & 0.61 & 0.54 & 0.11 \\
deepseek-r1:14b & 0.42 & 0.47 & 0.52 & 0.53 & 0.44 & 0.11 \\
deepseek-r1:7b & 0.17 & 0.21 & 0.21 & 0.23 & 0.19 & 0.06 \\
\bottomrule
\end{tabular}
\end{table*}

\begin{table}[!htb]
\caption{Accuracy on female and male names \emph{minus} accuracy on the letter-variable (XY) baseline, all 29 models, in percentage points (uncertain counts as incorrect). Positive${}={}$higher than XY. $\max|\Delta|$ is the larger absolute gap. Sorted by $\max|\Delta|$.}
\label{tab:gender_full}
\footnotesize
\centering
\begin{tabular}{@{}lccc@{}}
\toprule
\textbf{Model} & \textbf{Female} & \textbf{Male} & \textbf{max}\,$|\Delta|$ \\
\midrule
deepseek-r1:7b & $-$24.3 & $-$24.0 & 24.3 \\
llama3.2:3b & $-$4.8 & $-$4.5 & 4.8 \\
llama3.1:8b & $+$4.0 & $+$3.2 & 4.0 \\
deepseek-r1:32b & $-$3.8 & $-$1.4 & 3.8 \\
gemma3:12b & $-$3.6 & $-$3.0 & 3.6 \\
qwen3:14b & $-$2.7 & $-$3.1 & 3.1 \\
qwen3:1.7b & $+$2.9 & $+$3.0 & 3.0 \\
granite3.2:8b & $-$2.9 & $-$2.6 & 2.9 \\
gemma3:27b & $-$2.8 & $-$1.7 & 2.8 \\
gpt-5.4-mini & $+$0.5 & $-$2.7 & 2.7 \\
gemma4:e2b & $-$2.4 & $-$2.6 & 2.6 \\
gemma3:270m & $+$2.1 & $+$1.8 & 2.1 \\
gemma3:4b & $+$1.7 & $+$1.9 & 1.9 \\
Opus 4.7 & $+$1.3 & $+$1.9 & 1.9 \\
aya-expanse:32b & $-$1.6 & $-$1.8 & 1.8 \\
qwen3:8b & $-$1.7 & $-$1.3 & 1.7 \\
qwen3:0.6b & $+$1.6 & $+$1.7 & 1.7 \\
deepseek-r1:14b & $+$1.6 & $+$0.6 & 1.6 \\
gpt-oss:20 & $+$1.5 & $+$1.0 & 1.5 \\
gpt-oss:120b & $+$0.5 & $+$1.1 & 1.1 \\
gemma4:e4b & $+$0.1 & $+$1.0 & 1.0 \\
Sonnet 4.6 & $-$1.0 & $+$0.1 & 1.0 \\
gemma3:1b & $+$0.9 & $+$1.0 & 1.0 \\
qwen3:32b & $-$0.2 & $+$0.9 & 0.9 \\
qwen3:4b & $+$0.2 & $-$0.7 & 0.7 \\
mistral:7b & $+$0.6 & $+$0.3 & 0.6 \\
aya-expanse:8b & $+$0.6 & $+$0.4 & 0.6 \\
gemma4:12b & $-$0.0 & $-$0.5 & 0.5 \\
gemma4:31b & $+$0.0 & $+$0.2 & 0.2 \\
\bottomrule
\end{tabular}
\end{table}

\begin{table*}[!htb]
\caption{Accuracy by activity scenario, all 29 models (uncertain counts as incorrect). \textbf{Range} is the spread across the five activities. Sorted by Range.}
\label{tab:activity_full}
\footnotesize
\centering
\begin{tabular}{@{}lcccccc@{}}
\toprule
\textbf{Model} & \textbf{Party} & \textbf{Event} & \textbf{Gather} & \textbf{Meeting} & \textbf{Celebr.} & \textbf{Range} \\
\midrule
deepseek-r1:7b & 0.25 & 0.28 & 0.17 & 0.17 & 0.15 & 0.13 \\
qwen3:8b & 0.60 & 0.64 & 0.58 & 0.61 & 0.57 & 0.06 \\
llama3.1:8b & 0.43 & 0.49 & 0.49 & 0.44 & 0.46 & 0.06 \\
gpt-oss:20 & 0.77 & 0.73 & 0.72 & 0.75 & 0.71 & 0.06 \\
gemma3:27b & 0.65 & 0.67 & 0.63 & 0.66 & 0.62 & 0.05 \\
qwen3:32b & 0.70 & 0.74 & 0.71 & 0.73 & 0.70 & 0.04 \\
gemma4:e2b & 0.58 & 0.61 & 0.57 & 0.61 & 0.58 & 0.04 \\
gemma4:e4b & 0.63 & 0.62 & 0.59 & 0.59 & 0.60 & 0.04 \\
deepseek-r1:32b & 0.56 & 0.58 & 0.56 & 0.55 & 0.53 & 0.04 \\
llama3.2:3b & 0.46 & 0.48 & 0.47 & 0.48 & 0.45 & 0.04 \\
mistral:7b & 0.56 & 0.53 & 0.53 & 0.54 & 0.52 & 0.04 \\
deepseek-r1:14b & 0.49 & 0.49 & 0.46 & 0.47 & 0.46 & 0.04 \\
granite3.2:8b & 0.62 & 0.61 & 0.58 & 0.62 & 0.61 & 0.04 \\
qwen3:1.7b & 0.51 & 0.51 & 0.50 & 0.54 & 0.53 & 0.03 \\
gemma3:12b & 0.62 & 0.62 & 0.59 & 0.62 & 0.62 & 0.03 \\
gpt-5.4-mini & 0.64 & 0.66 & 0.66 & 0.67 & 0.65 & 0.03 \\
qwen3:4b & 0.56 & 0.57 & 0.57 & 0.58 & 0.56 & 0.03 \\
qwen3:0.6b & 0.54 & 0.54 & 0.55 & 0.56 & 0.55 & 0.03 \\
Opus 4.7 & 0.75 & 0.76 & 0.75 & 0.77 & 0.77 & 0.02 \\
aya-expanse:8b & 0.58 & 0.59 & 0.57 & 0.59 & 0.59 & 0.02 \\
gemma4:12b & 0.76 & 0.77 & 0.76 & 0.77 & 0.76 & 0.02 \\
gpt-oss:120b & 0.72 & 0.74 & 0.72 & 0.73 & 0.72 & 0.02 \\
gemma3:4b & 0.52 & 0.54 & 0.53 & 0.53 & 0.52 & 0.02 \\
aya-expanse:32b & 0.63 & 0.62 & 0.62 & 0.63 & 0.62 & 0.02 \\
gemma4:31b & 0.82 & 0.82 & 0.81 & 0.81 & 0.82 & 0.01 \\
qwen3:14b & 0.61 & 0.61 & 0.61 & 0.60 & 0.60 & 0.01 \\
Sonnet 4.6 & 0.75 & 0.75 & 0.75 & 0.76 & 0.76 & 0.01 \\
gemma3:270m & 0.50 & 0.50 & 0.50 & 0.49 & 0.50 & 0.01 \\
gemma3:1b & 0.53 & 0.53 & 0.53 & 0.54 & 0.54 & 0.01 \\
\bottomrule
\end{tabular}
\end{table*}

\begin{table*}[!htb]
\caption{Accuracy on each nationality group \emph{minus} accuracy on the abstract letter-variable (XY) baseline, all 29 models, in percentage points (uncertain counts as incorrect). Positive${}={}$higher on that nationality than on XY. Ind=Indian, Rus=Russian, Jpn=Japanese, Afr=African, Ger=German, Fre=French, Amr=American. $\max|\Delta|$ is the largest absolute gap. Sorted by $\max|\Delta|$.}
\label{tab:nationality_full}
\footnotesize
\centering
\begin{tabular}{@{}lcccccccc@{}}
\toprule
\textbf{Model} & \textbf{Ind} & \textbf{Rus} & \textbf{Jpn} & \textbf{Afr} & \textbf{Ger} & \textbf{Fre} & \textbf{Amr} & \textbf{max}\,$|\Delta|$ \\
\midrule
deepseek-r1:7b & $-$13.5 & $-$23.7 & $-$34.1 & $-$20.2 & $-$26.3 & $-$27.8 & $-$23.4 & 34.1 \\
deepseek-r1:32b & $-$1.9 & $-$0.5 & $-$1.6 & $-$2.2 & $-$0.8 & $-$9.5 & $-$1.7 & 9.5 \\
llama3.2:3b & $-$3.6 & $-$2.2 & $-$9.1 & $-$2.7 & $-$2.8 & $-$8.2 & $-$4.0 & 9.1 \\
llama3.1:8b & $-$2.2 & $+$1.9 & $+$3.2 & $+$7.5 & $+$3.8 & $+$4.8 & $+$6.2 & 7.5 \\
gemma3:12b & $-$2.1 & $-$3.6 & $-$3.9 & $-$3.8 & $-$2.2 & $-$4.8 & $-$2.8 & 4.8 \\
Opus 4.7 & $+$2.1 & $+$4.8 & $+$0.9 & $+$1.1 & $+$1.3 & $+$0.0 & $+$1.1 & 4.8 \\
qwen3:14b & $-$3.1 & $-$2.8 & $-$2.8 & $-$2.2 & $-$2.1 & $-$4.7 & $-$2.5 & 4.7 \\
qwen3:1.7b & $+$2.9 & $+$3.0 & $+$2.4 & $+$4.6 & $+$3.1 & $+$3.2 & $+$1.4 & 4.6 \\
gemma3:4b & $+$4.6 & $+$4.1 & $+$0.5 & $+$0.2 & $+$0.9 & $+$0.5 & $+$1.7 & 4.6 \\
gemma3:27b & $-$1.1 & $-$0.9 & $-$4.1 & $-$3.3 & $-$1.4 & $-$2.9 & $-$2.1 & 4.1 \\
deepseek-r1:14b & $+$2.5 & $+$3.6 & $-$2.8 & $+$3.9 & $+$0.5 & $-$2.2 & $+$2.1 & 3.9 \\
granite3.2:8b & $-$3.2 & $-$1.1 & $-$3.2 & $-$3.8 & $-$3.5 & $-$2.0 & $-$2.2 & 3.8 \\
gemma4:e2b & $-$2.6 & $-$2.4 & $-$2.9 & $-$3.5 & $-$2.2 & $-$1.6 & $-$2.5 & 3.5 \\
qwen3:0.6b & $+$0.3 & $+$3.3 & $+$2.5 & $-$0.5 & $+$1.5 & $+$2.1 & $+$2.2 & 3.3 \\
gpt-oss:20 & $+$1.4 & $+$0.5 & $+$0.1 & $+$1.9 & $+$1.5 & $+$0.3 & $+$3.3 & 3.3 \\
gpt-5.4-mini & $-$1.6 & $-$1.9 & $-$0.9 & $-$0.4 & $-$2.2 & $-$3.2 & $+$2.4 & 3.2 \\
gemma3:270m & $+$1.4 & $+$1.0 & $+$3.0 & $+$1.3 & $+$2.0 & $+$2.4 & $+$2.2 & 3.0 \\
qwen3:8b & $-$1.6 & $-$0.3 & $-$1.4 & $-$3.0 & $-$1.6 & $-$1.3 & $-$1.4 & 3.0 \\
gpt-oss:120b & $-$0.2 & $+$2.2 & $-$0.9 & $-$0.9 & $+$0.8 & $+$1.9 & $+$2.8 & 2.8 \\
aya-expanse:32b & $-$1.8 & $-$2.5 & $-$2.1 & $-$0.9 & $-$1.8 & $-$1.3 & $-$1.5 & 2.5 \\
gemma3:1b & $+$0.6 & $+$0.7 & $+$2.1 & $+$0.9 & $+$1.3 & $+$0.3 & $+$0.7 & 2.1 \\
qwen3:32b & $-$0.5 & $+$0.7 & $+$1.5 & $+$0.3 & $+$1.7 & $-$0.6 & $-$0.7 & 1.7 \\
mistral:7b & $-$0.6 & $+$1.7 & $+$1.7 & $+$0.3 & $+$1.1 & $+$0.1 & $-$1.1 & 1.7 \\
aya-expanse:8b & $+$0.4 & $+$1.4 & $+$0.4 & $+$0.5 & $+$1.7 & $-$0.8 & $+$0.2 & 1.7 \\
qwen3:4b & $+$0.2 & $-$1.5 & $+$0.4 & $-$0.2 & $+$0.3 & $-$0.6 & $-$0.3 & 1.5 \\
gemma4:12b & $+$0.9 & $+$0.0 & $-$0.7 & $+$0.9 & $-$1.2 & $-$1.4 & $-$0.5 & 1.4 \\
gemma4:e4b & $+$0.6 & $+$1.2 & $+$0.7 & $+$1.2 & $-$0.5 & $-$0.4 & $+$1.2 & 1.2 \\
Sonnet 4.6 & $-$0.1 & $+$0.6 & $-$0.7 & $-$0.6 & $-$1.1 & $-$0.6 & $-$0.2 & 1.1 \\
gemma4:31b & $+$0.4 & $-$0.2 & $-$0.3 & $-$0.1 & $+$0.5 & $-$0.2 & $+$0.5 & 0.5 \\
\bottomrule
\end{tabular}
\end{table*}


\begin{figure*}[t!]
\centering
\includegraphics[width=0.49\linewidth]{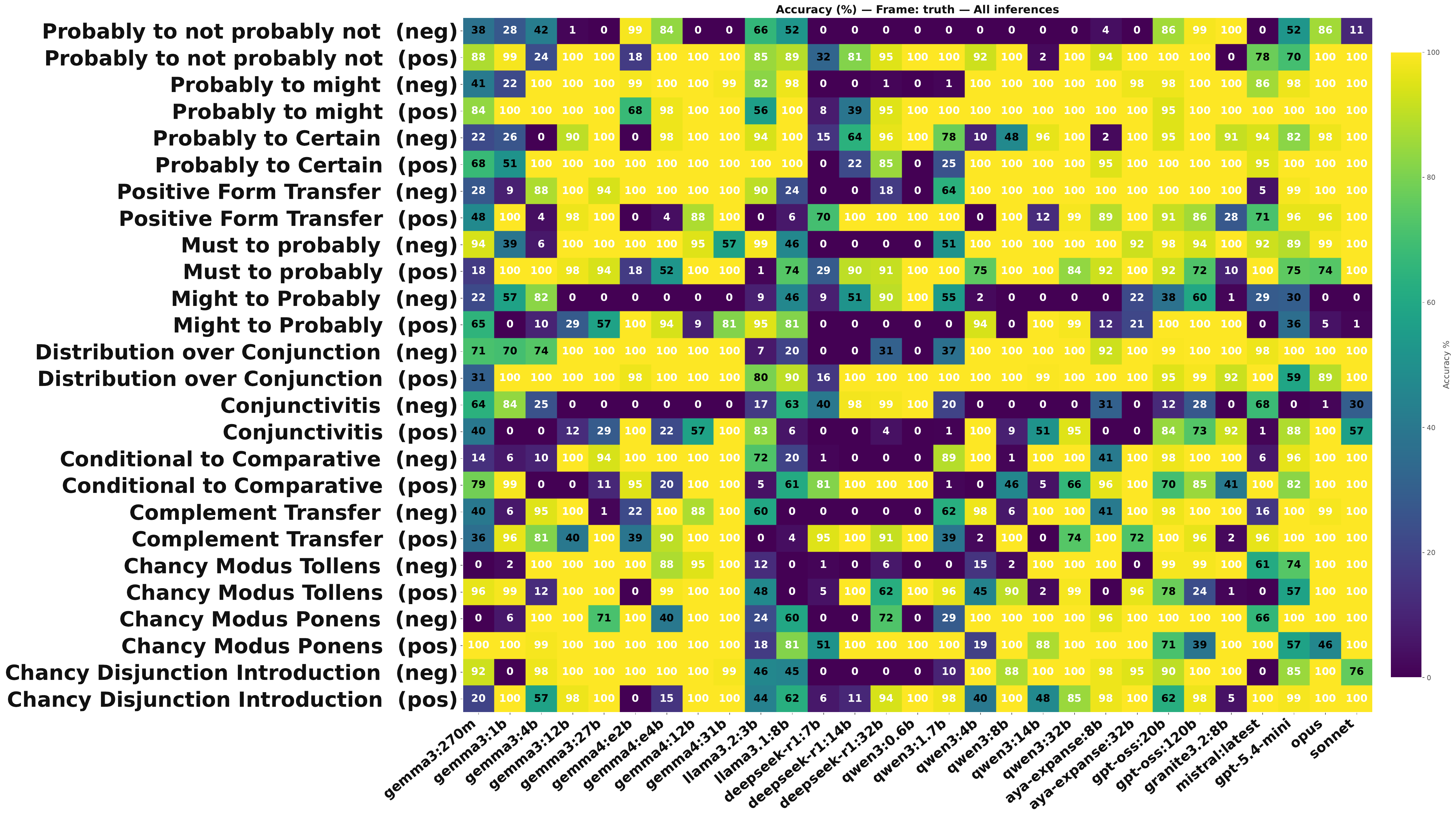}\hfill
\includegraphics[width=0.49\linewidth]{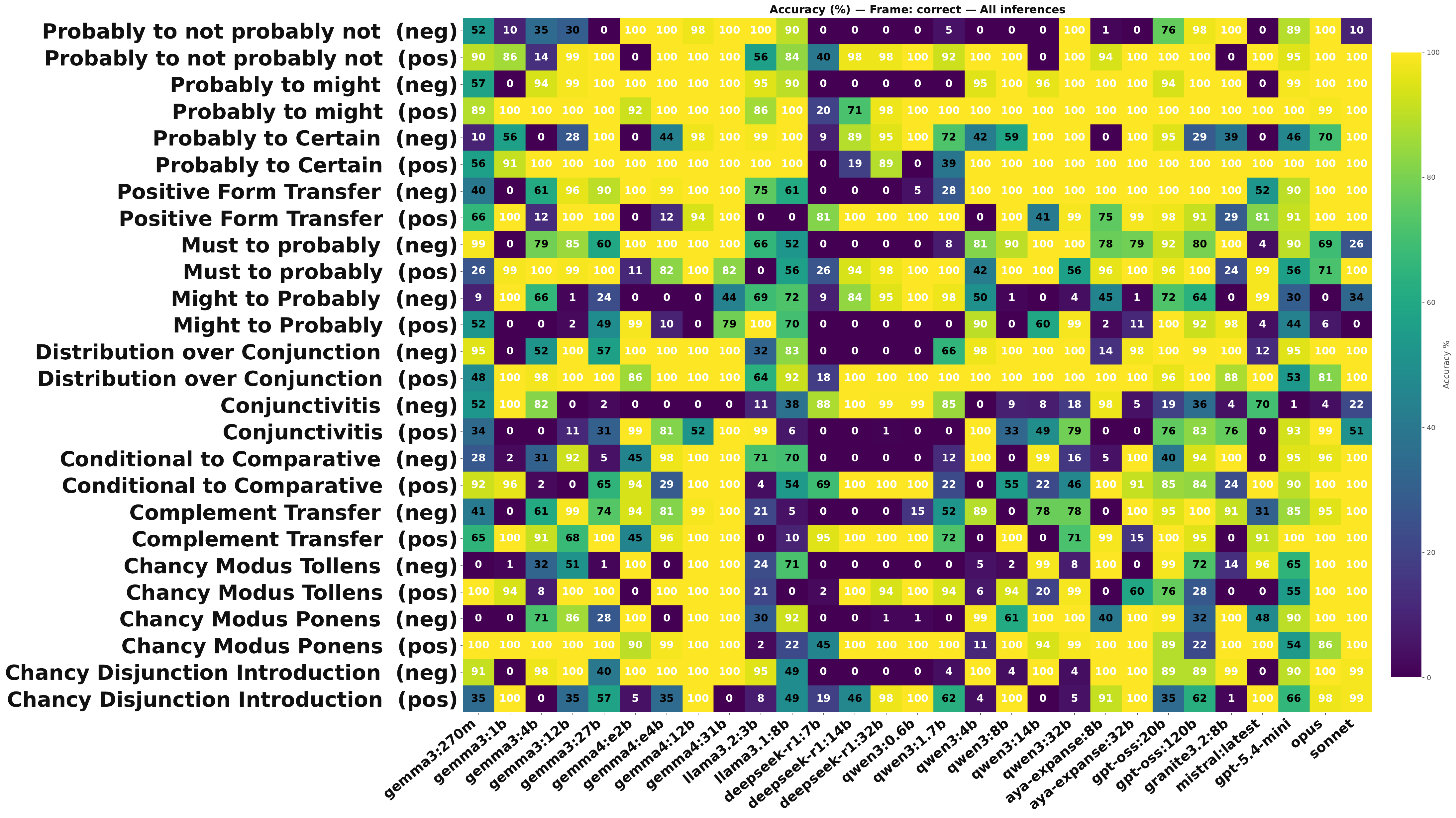}\\[1em]
\includegraphics[width=0.49\linewidth]{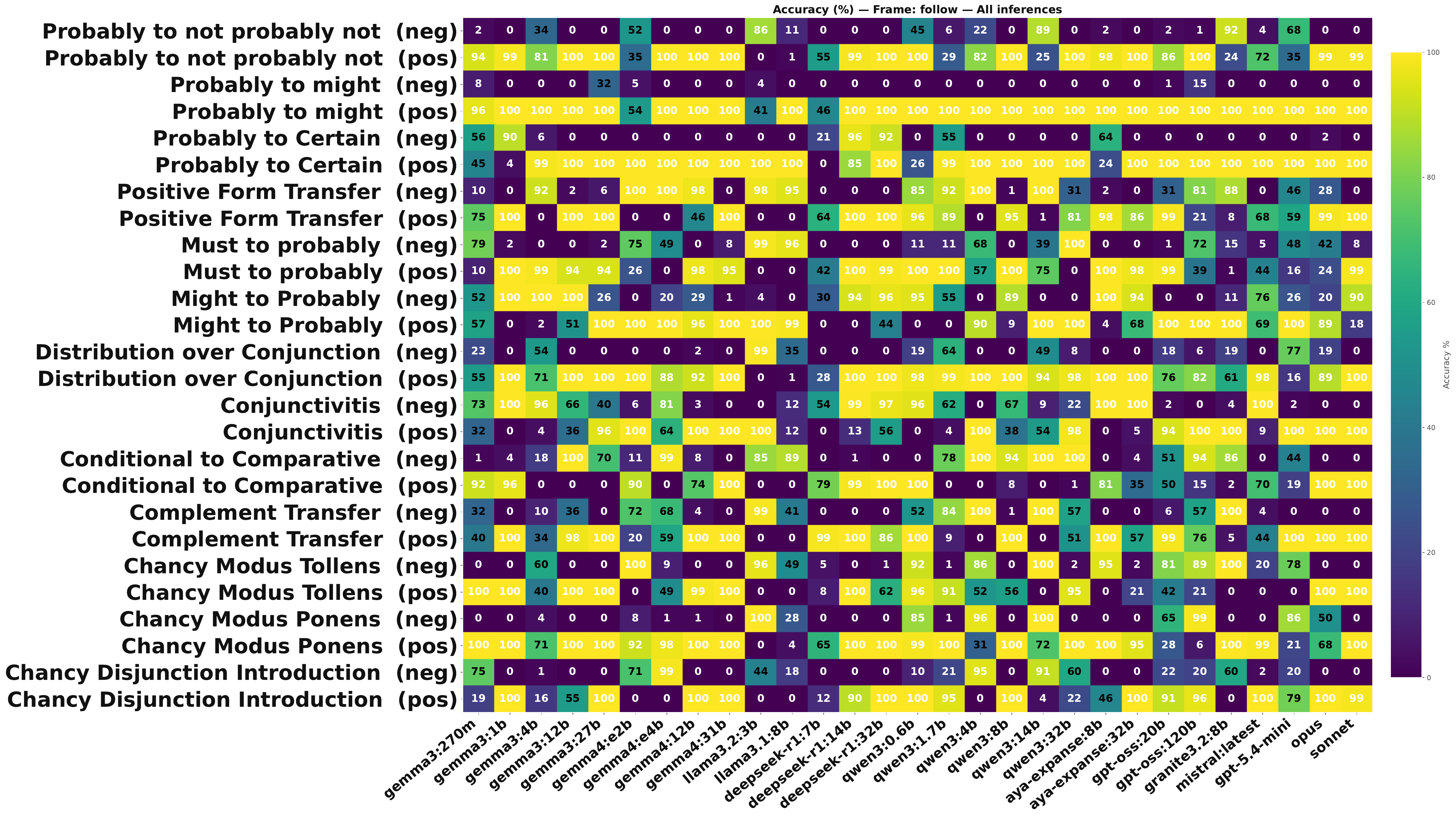}\hfill
\includegraphics[width=0.49\linewidth]{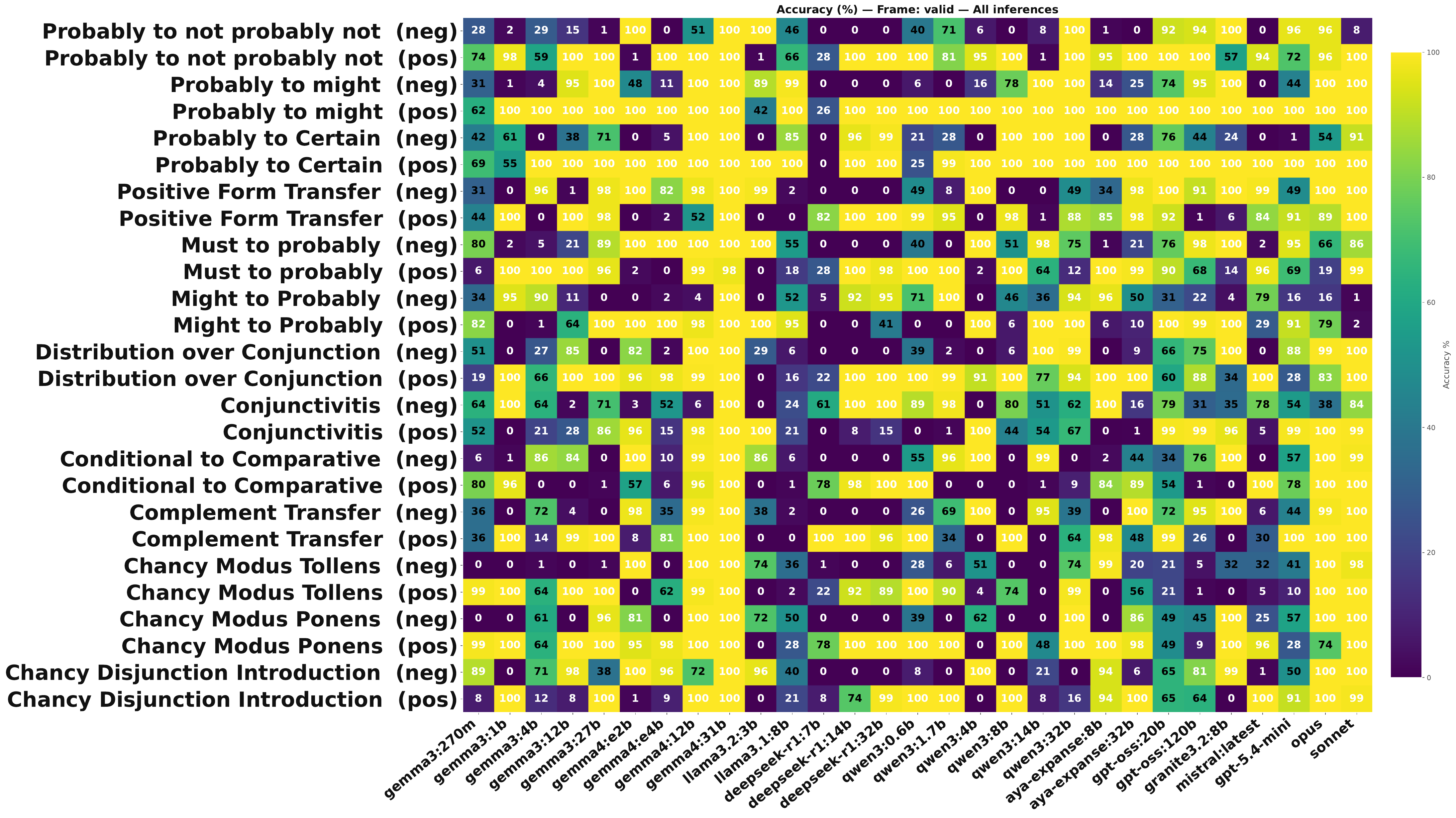}\\[1em]
\includegraphics[width=0.49\linewidth]{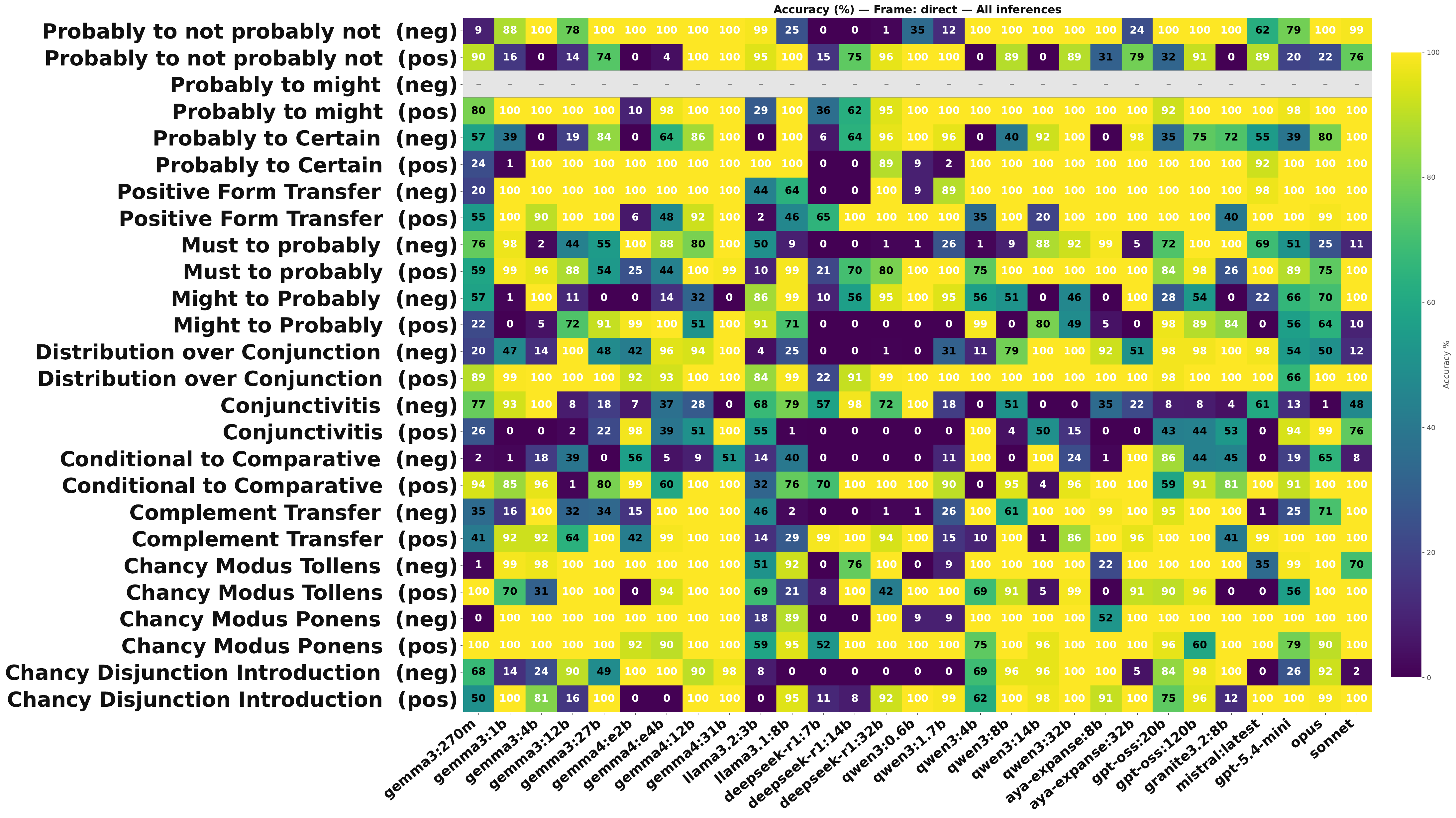}
\caption{Accuracy (\%) for every (template $\times$ negation) cell by model, shown separately for each of the five question forms. The same logical content swings from near-zero to near-ceiling across question forms, visualizing the question-form instability summarized in Table~\ref{tab:framesens}.}
\label{fig:heatmap}
\end{figure*}


\end{document}